\documentclass[10pt,twocolumn,twoside] {IEEEtran}
\usepackage{algorithm}
\usepackage{url}
\usepackage{cite}
\usepackage[usenames]{color}
\usepackage{amsfonts}
\usepackage{fancyhdr}
\usepackage{booktabs}
\usepackage{listings}%
\usepackage{arydshln}
\usepackage{amsmath,amssymb,amsthm}%
\usepackage{graphicx,psfrag,caption,subcaption}
\usepackage{listings}%
\usepackage{arydshln}
\usepackage{epstopdf}
\usepackage{graphicx,psfrag}
\input{mysymbol.sty}
\usepackage{algorithm}      
\usepackage[noend]{algpseudocode} 

\begin{document}

\title{Product Graph Learning from Multi-domain Data with Sparsity and Rank Constraints}

\author{Sai Kiran Kadambari,~\IEEEmembership{Student Member,} and Sundeep Prabhakar Chepuri,~\IEEEmembership{Member}  \\
        

\thanks{S.K. Kadambari and S.P. Chepuri are with the Department of Electrical Communication Engineering, Indian Institute of Science, Bangalore, India. E-mail: \{kadambarik; spchepuri\}@iisc.ac.in}
\thanks{This work is supported in part by the Pratiskha Trust Young Investigator Award and the SERB grant SRG/2019/000619. The precursor version has been published as \cite{kadambari2020learning}. }
}

\maketitle

\begin{abstract}
	In this paper, we focus on learning product graphs from multi-domain data. We assume that the product graph is formed by the Cartesian product of two smaller graphs, which we refer to as graph factors. We pose the product graph learning problem as the problem of estimating the graph factor Laplacian matrices. To capture local interactions in data, we seek sparse graph factors and assume a smoothness model for data. We propose an efficient iterative solver for learning sparse product graphs from data. We then extend this solver to infer multi-component graph factors with applications to product graph clustering by imposing rank constraints on the graph Laplacian matrices. Although working with smaller graph factors is computationally more attractive, not all graphs may readily admit an exact Cartesian product factorization. To this end, we propose efficient algorithms to approximate a graph by a  nearest Cartesian product of two smaller graphs. The efficacy of the developed framework is demonstrated using several numerical experiments on synthetic data and real data. 
\end{abstract}


\IEEEpeerreviewmaketitle

\section{Introduction} \label{sec:intro}

\IEEEPARstart{S}{ignal} processing and machine learning tools often leverage the underlying structure in data that may be available \emph{a priori} for solving inference tasks. Some examples of commonly used structures include low rank, sparsity, or network structures~\cite{sandryhaila2014big,shuman2013emerging}. 

Graphs offer mathematical tools to model complex interactions in modern datasets and to analyze and process them. Graphs occur naturally as sensing domains for data gathered from meteorological stations, road and internet traffic networks, and social and biological networks, to name a few. In these applications, the nodes of a graph act as indices of the data (or signals). The edges of a graph encode the pairwise relationship between the function values at the nodes. Hence such datasets are referred to as graph data (or signals). 

Taking this graph structure into account while processing data has well-documented merits for several signal processing and machine learning tasks like sampling and active learning~\cite{chen_2015_sampling,ortiz2019sparse,chepuri2017graph}, filtering~\cite{sandryhaila2014big}, and clustering~\cite{SpectralClusteringTutorial}, to list a few. These inference tasks within the fields of graph signal processing (GSP) and machine learning over graphs (GML) assume that the underlying graph is known or appropriately constructed depending on the application. For instance, in a sensor network with the sensors representing the nodes of a graph, the edge weight between any two sensors can be chosen as a function of its geographical proximity. 
Such a choice may not be useful in brain data processing or air quality monitoring applications, in which the similarity of the sensor measurements is not always related to the proximity of the sensors. 
Thus the underlying graph needs to be estimated using a data-driven approach when it is not available. The problem of inferring the underlying graph from data is referred to as \emph{graph learning}~\cite{dong2016learning,chepuri2016learning,kalofolias2016learn,egilmez2017graph,
dong19LearnGraphData,mateos19connectingdots,chepuri2018distributed,kumar2019structured,giannakis2018topology}.

\subsection{Related prior works} \label{subsec:priorwork}

The problem of inferring graph topologies from data is ill-posed. Therefore, a graph that best explains the data is often learned
by matching the properties of the data to the desired graph and exploiting any prior knowledge available about the desired graph. For an extensive overview on graph learning (also referred to as topology inference) techniques, see~\cite{dong19LearnGraphData} and \cite{mateos19connectingdots} (and references therein). 

Although a sample correlation or similarity graph (e.g., $k$-nearest neighbor graph or Gaussian similarity kernel)~\cite{SpectralClusteringTutorial} computed from data are simple graph learning methods, they are sensitive to noise or missing samples as they do not take into account any available prior information about the desired graph or data model. There are two commonly used data models when learning undirected graphs from data. The first model is a probabilistic graphical model of random graph data. In this data model, the assumption is that the graph structure is related to the inverse data covariance matrix (or the so-called precision matrix) that encodes conditional independence relationship of random variables indexed by the nodes. Graph learning, in this case, reduces to finding a maximum likelihood estimator of the inverse covariance matrix by modeling it as a regularized graph Laplacian matrix~\cite{lake2010discovering}. The second model is a smoothness model based on the quadratic total variation of data over the graph or sparsity in a basis related to the graph. Graph learning under the smoothness model amounts to finding a graph Laplacian matrix that promotes a smooth variation of the data over the desired graph~\cite{dong2016learning}. 

Typically we are interested in learning graphs that capture well the local interactions that many real-world datasets exhibit. Thus to obtain useful and meaningful graphs, we seek graph Laplacian matrices that are sparse with very few nonzero entries~\cite{dong2016learning,chepuri2016learning}. Next to a sparsity prior, it is also useful for graph-based clustering applications to infer graphs by incorporating spectral priors to obtain, for instance, multi-component or bipartite graphs~\cite{SpectralClusteringTutorial,nie2016constrained,kumar2019structured}.

In this work, we restrict our focus to a smoothness data model for structured graph learning, in which we consider a specific structure for the graph in the vertex domain, namely, the \emph{product structure}. Specifically, we focus on Cartesian product graphs with two smaller graph factors. Cartesian product graphs are useful to explain complex relationships in multi-domain graph data. For instance, consider a rating matrix in a recommender system such as Netflix or Amazon. It has a user dimension as well as an item or a movie dimension. The graph underlying such multi-domain data can often be factorized into a user graph and an item or a movie graph. Furthermore, the graph Laplacian matrix of a Cartesian product graph has a Kronecker sum structure with many interesting spectral properties that allow us to infer such graphs efficiently, or to perform graph filtering~\cite{sandryhaila2014big} and sampling~\cite{ortiz2019sparse} over product graphs efficiently.  

Assuming a probabilistic graphical data model,~\cite{kalaitzis2013Biglasso} presented Bigraphical Lasso (\texttt{BiGLasso}) \textemdash an algorithm for estimating precision matrices with a Kronecker sum structure. Although \texttt{BiGLasso} is not intended to estimate the Laplacian matrices of the graph factors that we are interested in this work, we can always project the inverse covariance matrices obtained from \texttt{BiGLasso} onto the space of valid Laplacian matrices to get a naive estimate of the Laplacian matrices of the graph factors (see Section~\ref{sec:numerical_expt} for more details). 

In a closely related work in~\cite{lodhi2020learning}, the authors also consider the problem of inferring product graphs as in the precursor~\cite{kadambari2020learning} of the paper, but without imposing sparsity or spectral constraints on the sought graph. The paper extends the precursor~\cite{kadambari2020learning} in various aspects, which we summarize in the next section as the main results.


\subsection{Main results and contributions} \label{subsec:mainresults}

This paper focuses on inferring undirected sparse Cartesian product graphs having two smaller undirected graph factors. We learn these graph factors by estimating the associated graph Laplacian matrices for the following cases.

\begin{itemize} 
\item We assume a smoothness data model and propose a convex optimization problem for estimating the Laplacian matrices of the graph factors. We present a generative model based on a Kronecker-structured factor analysis model. This model allows us to draw a connection between the smooth variation of data over the product graph and its graph factors. We develop an iterative algorithm to solve the proposed problem optimally by exploiting the symmetric structure of the Laplacian matrices. For a Cartesian product graph with $N$ nodes having graph factors with $P$ and $Q$ nodes such that $N = PQ$, the proposed algorithm incurs a significantly lower computational complexity of about order $P^2 + Q^2$ flops, whereas existing algorithms that do not leverage the Cartesian product structure costs about order $N^2$ flops.
\item For spectral clustering and subspace representation involving Cartesian product graphs, we extend the framework to learn Cartesian product graphs with multiple connected components by constraining the rank of the Laplacian matrices. For the resulting nonconvex optimization, we develop a solver based on cyclic minimization. Each subproblem of this cyclic minimization algorithm is solved optimally. As we see later, a $K$-component Cartesian product graph with $N$ nodes has $K_P$- and $K_Q$-component graph factors with $P$ and $Q$ nodes, respectively, such that $K = K_PK_Q$ and $N = PQ$. To estimate a $K$-component Cartesian product graph to perform graph clustering, the proposed algorithm incurs a computational complexity of about order $K_PP^2 + K_QQ^2$ flops that otherwise would cost about order $KN^2$ flops.

\item Working with smaller graph factors is computationally more attractive. However, not all graphs admit an exact Cartesian product factorization. In such cases, it is useful to approximate the graph of interest by a Cartesian product of two smaller graphs. To this end, we develop algorithms to find a {\it nearest} Kronecker sum factorization of a Laplacian matrix to compute the Laplacian matrices of the graph factors. We develop algorithms to obtain sparse and multi-connected graph factors as before.
 \end{itemize}
 
We demonstrate the usefulness of the developed algorithms through numerical experiments on synthetic and real datasets. The real datasets considered are related to air quality monitoring in India and object classification from multi-view images. 
  
\subsection{Notation and outline} \label{subsec:notation}
Throughout this paper, we will use upper and lower case boldface letters to denote matrices and column vectors. We will use calligraphic letters to denote sets. $|\cdot|$ denotes the cardinality of the set. $\mathbb{R}_+^N$ denotes the set of vectors of length $N$ with nonnegative real entries. $\mathbb{S}_+^N$ ($\mathbb{S}_{++}^N$) denotes the set of symmetric positive semidefinite (positive definite) matrix of size $N \times N$. $[\bbX]_{ij}$ and $x_i$ denote the $(i,j)$th element and $i$th element of the matrix $\bbX$ and vector $\bbx$, respectively. $\bbI_P $ denotes the identity matrix of size $P$. $\diag[\cdot]$ ($\bdiag[\cdot]$) is a (block) diagonal matrix with its argument along the main diagonal. 
${\rm tr}(\cdot)$ is the trace of the matrix. ${\rm vec}(\cdot)$ is the matrix vectorization operation. ${\rm vech}(\cdot)$ is the half-vectorization of a symmetric matrix obtained by vectorizing only the lower triangular part of the matrix. $\lambda_i(\cdot)$ denotes the $i$th smallest eigenvalue of its symmetric matrix argument. The symbols $\oplus$ and $\otimes$ represents the Kronecker sum and the Kronecker product, respectively. The symbol $\diamond$ denotes the Cartesian product between two graphs. $\bbx \succeq \bby$ denotes elementwise inequality between vectors $\bbx$ and $\bby$. The operator $\texttt{mat}(\bbx,M,N)$ returns the $M \times N$ matrix obtained by arranging entries of the vector $\bbx$ having $MN$ entries in column-major order.

We frequently use the following identities.
For matrices $\bbA$, $\bbB$, $\bbC$ and $\bbX$ of appropriate dimensions,
the equation $\bbA\bbX + \bbX\bbB = \bbC$ is equivalent to $(\bbA \oplus \bbB^T){\rm vec}(\bbX) ={\rm vec}(\bbC)$ and ${\rm vec}(\bbA\bbB\bbC) = (\bbC^T \otimes \bbA){\rm vec}(\bbB)$.
The Kronecker sum of two matrices $\bbA \in \reals^{M\times M}$ and $\bbB \in \reals^{N \times N}$ can be expressed as $\bbA \oplus \bbB = \bbI_N \otimes \bbA + \bbB \otimes \bbI_M$. 
The trace of the product of two matrices can be expressed as ${\rm tr}(\bbA^T\bbX) ={\rm vec}(\bbA)^T{\rm vec}(\bbX)$. Thus we have ${\rm tr}({\bbA}^T {\bbA}) = \| {\bbA}\|_F^2 ={\rm vec}({\bbA})^T{\rm vec}({\bbA})$. For an $N \times N$ symmetric and positive semidefinite matrix $\bbA$, we have~\cite[Proposition 1.3.4]{tao2012topics}
\begin{equation}
\sum\limits_{i=1}^K \lambda_i(\bbA) = \underset{\bbV \in \mathbb{R}^{N \times K}, \bbV^T\bbV = \bbI_{K}}{\rm minimize} \quad {\rm tr}({\bbV^T\bbA\bbV}). 
\label{eq:KyFan_knorm}
\end{equation}
We compactly denote the optimal solution $\bbV$ that contains as columns the $K$ eigenvectors corresponding to the $K$ smallest eigenvalues of $\bbA$ as $\bbV = \texttt{eigs}(\bbA, K)$.

The remainder of the paper is organized as follows. In Sections~\ref{sec:productgraphs} and~\ref{sec:productgraphsignal}, we discuss product graphs, subspace representation of product graphs, product graph signals, and a generative model for smooth product graph signals. In Section~\ref{sec:graphfromdata}, we develop algorithms for estimating graph factors from data. In Section~\ref{sec:factorization}, we present algorithms for finding a nearest Kronecker sum factorization for approximating a graph by a Cartesian product graph. We discuss the results from numerical experiments performed on synthetic and real datasets in Section~\ref{sec:numerical_expt}. The paper is concluded in Section~\ref{sec:conclusion}.

\section{Product graphs} \label{sec:productgraphs}
In this section, we give a brief introduction to Cartesian product graphs. We then describe how to exploit the Kronecker sum structure of a Cartesian product graph Laplacian matrix to efficiently compute low-dimensional spectral embeddings of the nodes of a product graph.

\subsection{Graph Laplacian matrix} \label{subsec:lapmatrix}

Consider a weighted undirected graph $\ccalG_{N} = (\ccalV_{N},\ccalE_{N})$ with $N$ nodes (or vertices). The sets $\ccalV_{N}$ and $\ccalE_N$ denote the vertex set and edge set, respectively. 
The structural connectivity of the graph $\ccalG_{N}$ is represented by the symmetric adjacency matrix $\bbW_N \in \reals^{N\times N}$. 
The $(i,j){\rm th}$ entry of $\bbW_N$ is positive if there is an edge between node $i$ and node $j$. 
We also assume that the graph $\ccalG_{N}$ has no self-loops. Thus the diagonal entries of $\bbW_N$ are all zero.
The diagonal node degree matrix of the graph $\ccalG_N$ is given by ${\rm diag}\left[\bbW_N{\bf1}_N\right] \in \reals^{N \times N}$.
The graph Laplacian matrix $\bbL_N \in\mathbb{S}_+^{N}$ associated with the graph $\ccalG_N$ is defined as $\bbL_N = {\rm diag}\left[\bbW_N{\bf1}_N\right] - \bbW_N$.

By construction the graph Laplacian matrix of an undirected graph is real, symmetric, and positive semidefinite. Thus ${\bf L}_N$ admits the following decomposition
\begin{equation}
\begin{aligned}
\bbL_{N} = \bbU_N \bbLam_N \bbU_N^T,
\end{aligned}
\end{equation}
where $\bbU_N \in \reals^{N\times N}$ is an orthonormal matrix that contains the eigenvectors~$\{\bbu_1,\cdots, \bbu_N\}$ as its columns and $\bbLam_N$ is a diagonal matrix that contains along its diagonal the eigenvalues $\{\lam_1(\bbL_N), \cdots,\lam_N(\bbL_N)\}$. We assume that the eigenvalues are ordered as $0 = \lambda_1(\bbL_N) \leq \lambda_2(\bbL_N) \leq \cdots \leq \lambda_N(\bbL_N)$. Furthermore, the all-one vector lies in null space of $\bbL_N$, i.e., $\bbu_1 = {\bf 1}_N$, without loss of generality. Thus we may denote the space of all the valid combinatorial Laplacian matrices of size $N$ by the set
\begin{equation}
\label{eq:Lapconstraint}
\ccalL_N := \{ \bbL \in \mathbb{S}_{+}^{N} \, | \, \bbL {\bf 1} = {\bf 0}, [\bbL]_{ij} = [\bbL]_{ji} \leq 0, \forall\,\, i \neq j\}.
\end{equation}
In this above set, the inequality $[\bbL]_{ii} \geq 0$ is implicit. Furthermore, the symmetric Laplacian matrix $\bbL_N$ can be compactly represented with the $M_N = N(N+1)/2$ nonduplicated elements $\bbl_N \in \mathbb{R}_+^{M_N}$ as
\begin{equation}
\label{eq:compact_L}
{\rm vec}(\bbL_N) = \bbD_N \bbl_N,
\end{equation}
where $\bbl_N \succeq {\bf 0}$ and the duplication matrix $\bbD_N$ of size $N^2 \times~M_N$ is explicitly expressed as
\begin{equation}
\label{eq:duplication_matrix}
\bbD_N^T = \sum_{i \geq j} {\boldsymbol \delta}_{ij} {\rm vec}^T({\boldsymbol \Theta}_{ij}).
\end{equation}
Here, ${\boldsymbol \delta}_{ij} \in \mathbb{R}^{M_N}$ is the canonical vector with 1 at position $(j-1)N + i - 0.5j(j-1)$, and zero elsewhere. The matrix ${\boldsymbol \Theta}_{ij} \in \mathbb{R}^{N \times N}$ is the canonical symmetric matrix with the value $-1$ in the position $(i,j)$, $1$ at the position $(i,i)$, and zero elsewhere, for all $i, j$.

\begin{figure}[!t]
\centering
\begin{subfigure}[t]{0.48\columnwidth}
\includegraphics[width=\columnwidth]{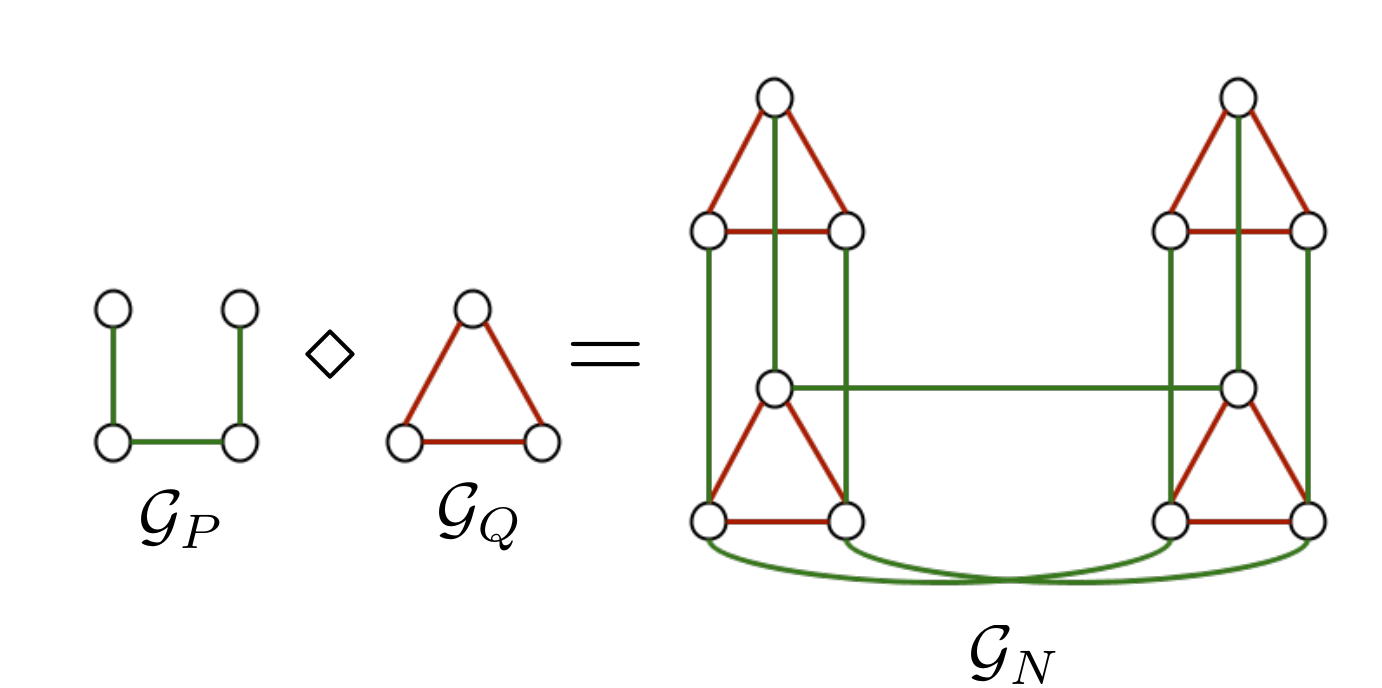}
\caption{}
\end{subfigure}
~
\begin{subfigure}[t]{0.48\columnwidth}
\includegraphics[width=\columnwidth]{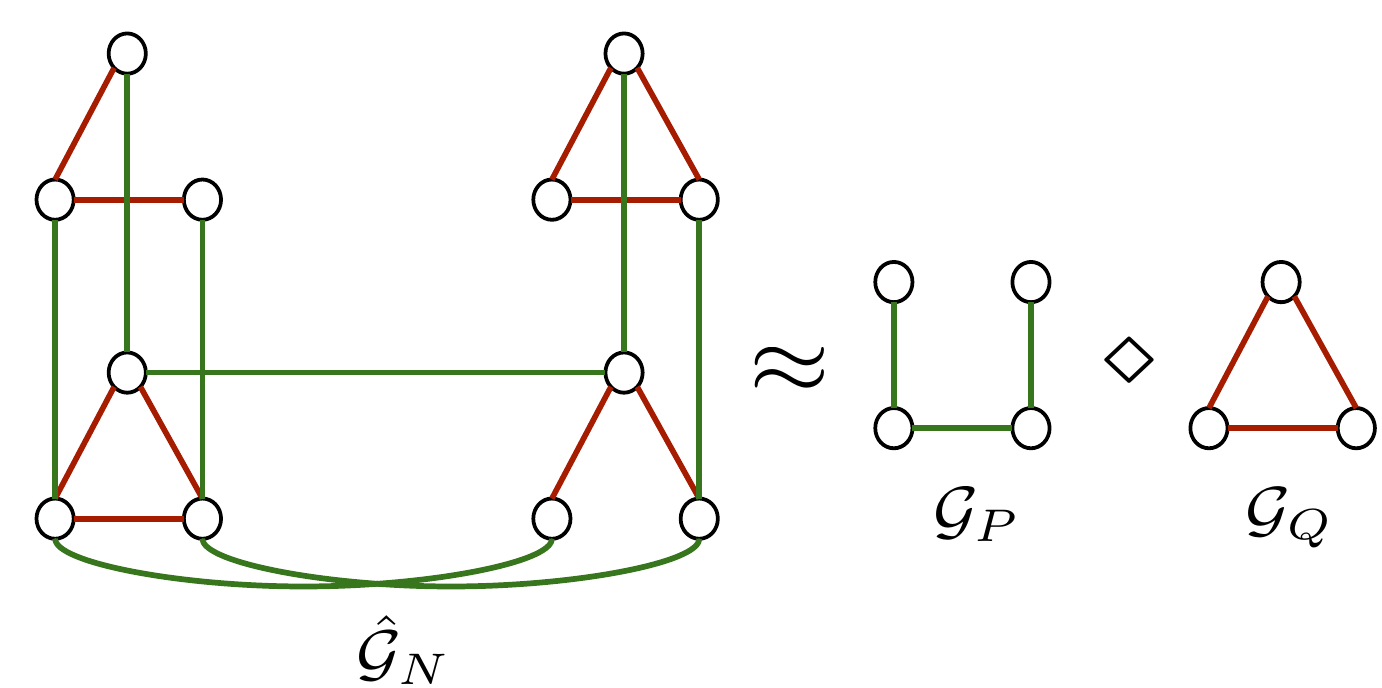}
\caption{}
\end{subfigure}
\caption{(a) Illustration of a Cartesian product graph $\ccalG_N$ with graph factors $\ccalG_P$ and $\ccalG_Q$. (b) A graph $\hat{\ccalG}_N$ is approximated by a Cartesian product of $\ccalG_P$ and $\ccalG_Q$.}
\label{fig:CartesianPG}
\end{figure}

\subsection{Product graph Laplacian matrix} \label{subsec:productgraphs}
Consider two graphs $\ccalG_P = (\ccalV_P,\ccalE_P)$ and $\ccalG_Q = (\ccalV_Q, \ccalE_Q)$ having $P$ and $Q$ nodes, respectively. Let $\bbL_P\in \mathbb{S}_{+}^{P}$ and $\bbL_Q \in \mathbb{S}_{+}^{Q}$ be the graph Laplacian matrices of $\ccalG_P$ and $\ccalG_Q$, respectively.
We assume that the graph $\ccalG_N$ can be factorized into two factors such that the Cartesian product of the graphs $\ccalG_P$ and $\ccalG_Q$ is $\ccalG_N$, that is,
\[
\ccalG_N = \ccalG_P \diamond \ccalG_Q \quad \text{with} \quad \ccalV_N = \ccalV_P \times \ccalV_Q.
\] 
In other words, the graphs $\ccalG_P$ and $\ccalG_Q$ form the graph factors of the Cartesian product graph $\ccalG_N$. The number of nodes in the Cartesian product graph $\ccalG_N$ is $N = P Q$. An illustration of a Cartesian product graph with $N=12$ nodes and graph factors having $P=3$ and $Q=4$ nodes is shown in Fig.~\ref{fig:CartesianPG}(a).

The graph Laplacian matrix $\bbL_N$ of $\ccalG_N$ can be expressed in terms of the Laplacian matrices of its graph factors $\ccalG_P$ and $\ccalG_Q$ as~\cite{sandryhaila2014big}
\begin{equation}\label{eq:CartesianProduct}
\begin{aligned}
\bbL_{N} = \bbL_{P} \oplus \bbL_{Q} = \bbI_Q \otimes \bbL_{P} +  \bbL_{Q} \otimes \bbI_P. 
\end{aligned}
\end{equation}
Similar to \eqref{eq:compact_L}, we can compactly represent $\bbL_P$ and $\bbL_Q$ as 
\begin{equation}
{\rm vec}(\bbL_P) = \bbD_P \bbl_P \quad \text{and} \quad {\rm vec}(\bbL_Q) = \bbD_Q \bbl_Q,
\label{eq:compact_LpLq}
\end{equation}
where $\bbl_P \in \mathbb{R}^{M_P}_+$ with $M_P = P(P+1)/2$ and $\bbl_Q \in \mathbb{R}^{M_Q}_+$ with $M_Q = Q(Q+1)/2$. Here, $\bbD_P \in \mathbb{R}^{P^2 \times M_P}$ and $\bbD_Q \in \mathbb{R}^{Q^2 \times M_Q}$ are the duplication matrices defined in~\eqref{eq:duplication_matrix}. We later on use the fact that the matrices $\bbD_P^T\bbD_P \in \mathbb{S}_{++}^{M_P}$ and $\bbD_Q^T\bbD_Q \in \mathbb{S}_{++}^{M_Q}$ are diagonal.

%
Let us denote the eigenvalue decomposition of the Laplacian matrices $\bbL_{P}$ and $\bbL_{Q}$ as $\bbL_P = \bbU_P\bbLam_P\bbU_P^T$ and $\bbL_Q = \bbU_Q\bbLam_Q \bbU_Q^T$, respectively. Here, $\bbU_P \in \reals^{P\times P}$ and $\bbU_Q\in \reals^{Q \times Q}$ are the eigenvector matrices, and $\bbLam_P \in \reals^{P\times P} $ and $\bbLam_Q \in \reals^{Q\times Q}$ are the matrices of eigenvalues.
Then the eigenvalue decomposition of $\bbL_N$ can be expressed as~\cite{horn1994topics}
\begin{equation}\label{eq:EVDCartesian} 
\begin{aligned}
\bbL_N &= (\bbU_P\bbLam_P\bbU_P^T) \oplus (\bbU_Q\bbLam_Q\bbU_Q^T)\\
&= (\bbU_Q \otimes \bbU_P)(\bbLam_P \oplus \bbLam_Q)(\bbU_Q\otimes\bbU_P)^T.
\end{aligned}
\end{equation}
This means that, the eigenvector matrix has a Kronecker product structure $\bbU_N = {\bf U}_Q \otimes {\bf U}_P$ and
the eigenvalue matrix has a Kronecker sum structure $\bbLam_N = {\boldsymbol \Lambda}_P \oplus {\boldsymbol \Lambda}_Q$. This Kronecker structure can be exploited to efficiently perform clustering on product graphs as discussed next.

\subsection{Subspace representation for product graphs} \label{subsec:subspace_graphs}

The algebraic multiplicity of the zero eigenvalue of a graph Laplacian matrix determines the number of connected components of the graph~\cite{spielman2012spectral}. This means that the rank of the graph Laplacian matrix of a $K$-connected graph with $N$ nodes is $N-K$.
The matrix containing the $K$ eigenvectors associated with the $K$ smallest eigenvalues of the graph Laplacian yields a low-dimensional node representation (i.e., a $K$-dimensional node embedding), which preserves the information about the connectivity of the nodes in the graph. In particular, the $n$th row of this matrix forms the coordinates of the $n$th node in the low-dimensional subspace and is used for clustering the nodes of the graph~\cite{ng2002spectral}.

A direct consequence of the Kronecker sum structure of the eigenvalue matrix in \eqref{eq:EVDCartesian} is that a $K$-component product graph can be equivalently represented as the Cartesian product of two smaller graph factors having $K_P$ and $K_Q$ components with $K = K_PK_Q$. Suppose the graph $\ccalG_N = \ccalG_P \diamond \ccalG_Q$ has $K$ \emph{connected components}. Then we have ${\rm rank}(\bbL_N) = N-K$, ${\rm rank}(\bbL_P) = P-K_P = R_P$, ${\rm rank}(\bbL_Q) = Q-K_Q = R_Q$. Further, from \eqref{eq:EVDCartesian}, the $K_P$-dimensional and $K_Q$-dimensional node embeddings can be used to efficiently compute the $K$-dimensional representation for $\ccalG_N$. This is particularly useful in reducing the computational costs incurred while clustering large graphs.

\section{Product graph signals} \label{sec:productgraphsignal}

This section discusses product graph signals and provides a generative model for smooth signals on a product graph.  

Let ${\bf x}_i \in \mathbb{R}^{N}$ be a signal defined on the product graph $\ccalG_N$ with its $i$-th entry indexed by the $i$th node of $\ccalG_N$. Let us collect a set of $T$ such graph signals $\{\bbx_i\}_{i = 1}^T$ in a matrix $\bbX \in \reals^{N\times T}$. 
Since every node in $\ccalG_N$ can be indexed by a pair of vertices on its graph factors, we can reshape $\bbx_i$ as a matrix $\bbX_i = {\rm mat}(\bbx_i, P,Q) \in \reals^{P\times Q}$, i.e., $\bbx_i ={\rm vec}(\bbX_i)$, for $i = 1,\ldots,T$.
This means that each $\bbx_i$ represents a multi-domain graph data with the entries of $\bbX_i$ indexed by the nodes of the graph factors $\ccalG_P$ and $\ccalG_Q$. 

\subsection{Smoothness data model}
A smoothness metric based on the quadratic total variation with respect to the underlying graph Laplacian matrix is often used to quantify how well the signal $\bbx_i$ is related to the supporting graph. Specifically, we define the metric 
\[
\bbx_i^T \bbL_N \bbx_i = \sum_{(m,n) \in \ccalE_N} [\bbW_N]_{mn} ([\bbx_i]_m - [\bbx_i]_n)^2,
\]
which penalizes signals $\bbx_i$ for which neighboring nodes 
have very different values. Clearly, for constant signals, this quadratic term is zero. For Cartesian product graphs [cf. \eqref{eq:CartesianProduct}], this smoothness promoting quadratic term can be explicitly expressed in terms of its graph factors as
\begin{equation}
\label{eq:TraceExpansion}
\begin{aligned}
\bbx_i^T(\bbL_P \oplus \bbL_Q)\bbx_i  
&={{\rm vec}}^T(\bbX_i){\rm vec}(\bbL_P \bbX_i + \bbX_i \bbL_Q) \\ 
&={\rm tr}(\bbX_i^T \bbL_P \bbX_i) + {\rm tr}(\bbX_i \bbL_Q \bbX_i^T).\end{aligned}
\end{equation}
This means that the quadratic norm induced by the Cartesian product graph is separable and can be simplified as the sum of the quadratic total variation of the signals collected in the rows and columns of $\bbX_i$ with respect to its graph factors 
$\ccalG_P$ and $\ccalG_Q$. For $T$ snapshots of the data, this quadratic norm generalizes as
\begin{eqnarray}
\sum\limits_{i=1}^T\bbx_i^T(\bbL_P \oplus \bbL_Q)\bbx_i  
&=& \sum\limits_{i=1}^T {\rm tr}(\bbX_i^T \bbL_P \bbX_i) + {\rm tr}(\bbX_i \bbL_Q \bbX_i^T) \nonumber \\ 
&=& {\rm tr}(\bbL_P\bbS_P) + {\rm tr}(\bbL_Q\bbS_Q), 
\label{eq:TraceExpansion2}
\end{eqnarray}
where we have defined the sample data covariance matrices $\bbS_P = \sum\limits_{i=1}^T \bbX_i\bbX_i^T \in \mathbb{R}^{P \times P}$ and $\bbS_Q = \sum\limits_{i=1}^T \bbX_i^T\bbX_i \in \mathbb{R}^{Q \times Q}$.

\subsection{Kronecker-structured factor analysis model}\label{sec:Kron_FA}
Next, we describe a generative model for the multi-domain data that vary smoothly across the edges of a product graph in terms of its graph factors. The presented modeling is an extension of the factor analysis model for single domain datasets~\cite{dong2016learning} to multi-domain data. 

Since the eigenvector matrix $\bbU_N$ has a Kronecker product structure, we can synthesize $\bbx_i$ as
\begin{equation}\label{eq:GFT}
\begin{aligned}
{\bbx}_i = \bbU_N\tilde{\bbx}_i = (\bbU_Q \otimes \bbU_P)\tilde{\bbx}_i ={\rm vec}(\bbU_P\tilde{\bbX}_i \bbU_Q^T),
\end{aligned}
\end{equation}
where $\tilde{\bbx}_i ={\rm vec}(\tilde{\bbX}_i)$. Let us consider a noisy version of ${\bf X}_i$ given by
\begin{equation}\label{eq:jointfac}
\begin{aligned}
\bbY_i = {\bf X}_i + \bbE_i = \bbU_P \tilde{\bf X}_i \bbU_Q^T + \bbE_i,
\end{aligned}
\end{equation}
where $\bbY_i \in \reals^{P \times Q}$ is the $i$th snapshot of the multi-domain noisy graph data, ${\bf E}_i \in \mathbb{R}^{P \times Q}$ is the observation noise, and $\tilde{\bf X}_i \in \reals^{P \times Q}$ is the latent factor matrix with loading matrices $\bbU_P$ and $\bbU_Q$, which {\it jointly} influence the observations $\bbY_i$. 

On vectorizing \eqref{eq:jointfac}, we get 
\begin{equation}\label{eq:FactorAnalysisModel}
\begin{aligned}
\bby_i = (\bbU_Q \otimes \bbU_P)\tilde{\bf x}_i + {\bf e}_i,
\end{aligned}
\end{equation}
where $\bby_i \in \reals^{PQ}$ with $\bby_i ={\rm vec}(\bbY_i)$. 
We assume that ${\bf e}_i ={\rm vec}({\bf E}_i)$ is distributed as ${\bf e}_i \sim \ccalN({\bf 0}, \sigma^2\bbI_N)$. We refer to this model as the Kronecker-structured factor analysis model.

Next, we define a Gaussian prior distribution $p(\tilde{\bf x}_i)$ over the latent factors along with a Gaussian conditional distribution $p({\bf y}_i | \tilde{\bf x}_i)$ for the observed variable ${\bf y}_i$ conditioned on the value of the latent variable $ \tilde{\bf x}_i$. More specifically, the prior distribution over $\tilde{\bf x}_i$ is given by
\begin{equation}\label{eq:Prior}
p(\tilde{\bf x}_i) =  \ccalN({\bf 0}, [\bbLam_P \oplus \bbLam_Q]^\dagger),
\end{equation}
where $[\bbLam_P\oplus\bbLam_Q]^\dagger \in \mathbb{R}^{N \times N}$ is a diagonal matrix and the conditional distribution of ${\bf y}_i$ conditioned on $\tilde{\bf x}_i$ is given by
\begin{equation*}\label{eq:ProbDistru_1}
p({\bf y}_i | \tilde{\bf x}_i) = \ccalN([\bbU_Q\otimes\bbU_P]\tilde{\bf x}_i, \sigma^2\bbI_N).
 \end{equation*}
Then the maximum a posteriori (MAP) estimate of $\tilde{\bbx}_i$ is
\begin{align}
& \arg \max_{\tilde{\bbx}_i} \, \ln p({\bf y}_i | \tilde{\bf x}_i) + \ln p(\tilde{\bbx}_i) \label{eq:MAPestimate} \\
&= \arg \min_{\tilde{\bbx}_i} \,\, \alpha \|\bby_i - \left(\bbU_Q\otimes\bbU_P\right) \tilde{\bbx}_i\|_2^2 
 + \, \tilde{\bbx}_i^T (\bbLam_P\oplus\bbLam_Q)\tilde{\bbx}_i, \nonumber
 \end{align}
where the constant $\alpha >0$ is related to the noise variance $\sigma^2$. 
Using~\eqref{eq:GFT} in \eqref{eq:MAPestimate} and from \eqref{eq:TraceExpansion}, the MAP estimate of ${\bbX}_i$ is obtained by solving
\begin{align*}
\arg \min_{{\bbX}_i}\quad \alpha\|\bbY_i - {\bbX}_i\|_F^2 +  {\rm tr}(\bbX_i^T \bbL_P \bbX_i) + {\rm tr}(\bbX_i \bbL_Q \bbX_i^T).
 \end{align*}
This means that the data that follows the Kronecker-structured factor analysis model \eqref{eq:jointfac} with a Gaussian prior on the latent factors as in \eqref{eq:Prior} yield graph data that are {\it simulatenously smooth} on the graph factors of the Cartesian product graph. More generally, we can jointly denoise $T$ snapshots of the noisy data $\bbY_i$, for $i=1,\ldots,T$, using the graph regularizer in \eqref{eq:TraceExpansion2}, as 
\begin{align}
\underset{{{\bbX}_i, i=1,\ldots, T}}{\rm minimize} \quad \sum\limits_{i=1}^T\alpha\|\bbY_i - {\bbX}_i\|_F^2 &+  {\rm tr}(\bbX_i^T \bbL_P \bbX_i) \nonumber \\ 
& \>  + {\rm tr}(\bbX_i \bbL_Q \bbX_i^T). \label{eq:dualgraphreg}
 \end{align}
 
In what follows, based on \eqref{eq:dualgraphreg}, we formulate product graph learning as the problem of estimating the Laplacian matrices of the graph factors underlying the available data.

\section{Computing graph factors from data}\label{sec:graphfromdata}

In this section, we present solvers for the problem of learning the graph factors underlying data with the assumption that the data is smooth with respect to the product graph. As discussed in Section~\ref{sec:intro}, there are several existing algorithms that estimate the graph Laplacian matrix from the available data. These methods ignore the fact that the underlying graph $\ccalG_N$ can be factorized as the Cartesian product of $\ccalG_P$ and $\ccalG_Q$. When this additional information is accounted for, we get computationally efficient algorithms. To begin with, we assume that the observed data is noiseless, i.e., $\bbY_i = \bbX_i$ for $i=1,2,\ldots,T$ (see the discussion in Section~\ref{sec:numerical_expt} when the observed data is noisy or incomplete) and solve the problem of learning {\it sparse} graph factors that sufficiently explain the data. Then, we extend this solution to obtain graph factors with multiple connected components by introducing {\it rank} constraints on the graph factor Laplacian matrices.

\subsection{Learning sparse graph factors}\label{sec:PGLprob}
%
We jointly estimate $\bbL_P$ and $\bbL_Q$ by restricting our search to the set of Laplacian matrices defined in \eqref{eq:Lapconstraint}. That is, we propose to solve the following \emph{product graph learning} (\texttt{PGL}) problem:
\begin{align}
&\underset{\bbL_P \in \ccalL_P, \,\, \bbL_Q \in \ccalL_Q}{{\rm minimize}} \quad
 {\rm tr}(\bbL_P \bbS_P) + {\rm tr}( \bbL_Q \bbS_Q) +  h(\bbL_P,\bbL_Q) \nonumber \\
&\quad \text{subject to} \quad \quad {\rm tr}(\bbL_P) = P \quad \text{and} \quad {\rm tr}(\bbL_Q)= Q, 
\tag{$\mathcal{P}1$}
\label{eq:PGL_modeling}
\end{align}
where the trace equality constraints avoid the trivial solution. It can be shown that ${\rm tr}(\bbL_i) = 2 \| {\rm vec}(\bbW_i)\|_1$, which is a commonly used convex approximation of the sparsity-inducing $\ell_0$-norm penalty. 
Since constraining $\| {\rm vec}(\bbW_i)\|_1$ for $i\in\{P,Q\}$, in this case, only changes the scale of the solution, we propose to choose the penalty term
\[
 h(\bbL_P,\bbL_Q) =\beta_1 \|\bbL_P\|_F^2+ \beta_2 \|\bbL_Q\|_F^2
\]
with parameters $\beta_1 > 0$ and $\beta_2 > 0$ in the objective to control the sparsity of $\ccalG_P$ and $\ccalG_Q$, respectively, by controlling the distribution of the edge weights. See~\cite{kalofolias2016learn} for other regularizers that may be used to obtain sparse graphs. Given $\bbS_P$ and $\bbS_Q$, the optimization problem \eqref{eq:PGL_modeling} is convex in $\bbL_P$ and $\bbL_Q$. 

Now, we present a very efficient solver for~\eqref{eq:PGL_modeling}, by exploiting the symmetric structure of the Laplacian matrices and by compactly representing it using \eqref{eq:compact_LpLq}. Let us define $\bbl \in \mathbb{R}_+^{M}$ with $M = M_P + M_Q$ as $\bbl = [\bbl_P^T, \bbl_Q^T]^T$. 
Using the identities  
\[
{\rm tr}(\bbL_i\bbS_i) = {\rm vec}^T(\bbS_i)\bbD_i\bbl_i 
\]
and 
\[
\| \bbL_i \|_F^2 = {\rm tr}(\bbL_i^2) = \bbl_i^T \bbD_i^T\bbD_i\bbl_i 
\]
for $i \in \{P,Q\}$, the optimization problem~\eqref{eq:PGL_modeling} can equivalently be written as
\begin{align}
&\underset{\bbl \in \mathbb{R}^M}{\rm minimize} \quad  
\frac{1}{2} \, \bbl^T {\rm diag}(\bbp_{\rm d}) \bbl \, + \, \bbq^T_{\rm d} \bbl \nonumber \\
& \text{subject to } \quad \bbC\bbl = \bbd \quad \text{and} \quad \bbl \succeq {\bf 0}
\label{eq:QP0}
\end{align}
with known parameters ${\rm diag}(\bbp_{\rm d}) \in \mathbb{S}_{++}^M$, $\bbq_{\rm d} \in \mathbb{R}^M$, $\bbC \in~\mathbb{R}^{L \times M}$, and $\bbd \in \mathbb{R}^{L}$, where $L = P+Q+2$. Here, 
\begin{align}
&{\rm diag}(\bbp_{\rm d}) := {\rm bdiag}[ {\rm diag}(\bbp_{{\rm d},P}), {\rm diag}(\bbp_{{\rm d},Q})] \nonumber \\ 
&\quad\quad\quad\quad = {\rm bdiag}[2\beta_1\bbD_P^T\bbD_P, \, 2\beta_2\bbD_Q^T\bbD_Q]; \label{eq:p0} 
\end{align}
and  
\begin{align}
&\bbq^T_{\rm d} :=[\bbq_{{\rm d},P}^T, \bbq_{{\rm d},Q}^T]= [{\rm vec}^T({\bbS}_P)\bbD_P, {\rm vec}^T({\bbS}_Q)\bbD_Q]. \label{eq:q0}
\end{align}
The subscript ``${\rm d}$" represents parameters related to the data. 
The parameters related to the equality constraints are [see Appendix~\ref{sec:lap_constraints}]
\begin{equation}
\bbC = {\rm bdiag}[\bbC_P,\bbC_Q] \quad \text{and}\quad \bbd = \left[\bbd_P^T, \bbd_Q^T \right]^T
\label{eq:C_and_d}
\end{equation}
with
\begin{equation*}
\bbC_i = \left[\begin{array}{c}{\rm vec}^T({\bbI}_i)\bbD_i \\({\bf 1}_i^T \otimes \bbI_i)\bbD_i\end{array}\right] \quad \text{and}\quad   
\bbd_i = \left[i, {\bf 0}_i^T\right]^T
\end{equation*}
for $i \in \{P,Q\}$. 
\begin{algorithm}[!t]
\caption{Product graph learning}\label{alg:laplacian_estimator}
\begin{algorithmic}[1]
\Function{\texttt{PGL}}{$\bbp$, $\bbq$, $\bbC$, $\bbd$, $\texttt{Tol}$, $\rho$}

\State Initialize $\bbmu \gets {\bf 0}$ 
\While{ $\|\bbC\bbl - \bbd\|_2 < \texttt{Tol}$} 
\State $\bbl \gets \left\{{\rm diag}^{-1}(\bbp)[\bbC^T\bbmu - \bbq] \right\}_+$
\State $\bbmu \gets \bbmu - \rho [ \bbC\bbl - \bbd]$
\EndWhile

\Return $\bbl$
\EndFunction
\end{algorithmic}
\end{algorithm}

The optimization problem~\eqref{eq:QP0} is a special case of a quadratic program in which the matrix associated with the quadratic term in the objective function is diagonal. This optimization problem can be solved efficiently and optimally using Algorithm~\ref{alg:laplacian_estimator}, which is developed in Appendix~\ref{sec:qpsolver}. The optimal solution to~\eqref{eq:PGL_modeling} is given by the iterative procedure 
\[
\bbl^\star = \texttt{PGL}(\bbp_{\rm d}, \bbq_{\rm d}, \bbC, \bbd, \texttt{Tol}, \rho). 
\]
Although we concatenate $\bbl_P$ and $\bbl_Q$, and present a solver for $\bbl$ as described above, it is computationally less expensive to solve for $\bbl_P$ and $\bbl_Q$ separately as it is easy to see that the problem is separable in these variables. Specifically, the solution to \eqref{eq:PGL_modeling} is given by $\bbl_i^\star = \texttt{PGL}(\bbp_{{\rm d},i}, \bbq_{{\rm d},i}, \bbC_i, \bbd_i, \texttt{Tol}, \rho)$ for $i \in \{P,Q\}$. By doing so, we incur a per iteration computational cost of the order $P^2 + Q^2$ flops (cf. Appendix~\ref{sec:qpsolver}) and not order $(P+Q)^2$ flops. 

In the next section, we specialize the problem of learning sparse graph factors discussed in this section to the specific case in which the graphs $\ccalG_P$ and $\ccalG_Q$ each have multiple connected components.

\subsection{Learning graph factors with rank constraints}\label{sec:RCPGL_prob}

As discussed in Section~\ref{subsec:subspace_graphs}, if the underlying graph has a product structure, then instead of learning a large multi-component graph with $N$ nodes, we may learn smaller multi-component graph factors with $P$ and $Q$ nodes, or learn the low-dimensional embeddings of the nodes in $\ccalG_N$, by computing the low-dimensional representations for $\ccalG_P$ and~$\ccalG_Q$.

Suppose that the graphs $\ccalG_P$ and $\ccalG_Q$ have $K_P$ and $K_Q$ connected components, respectively. Although tuning $\beta_1$ and $\beta_2$ in \eqref{eq:PGL_modeling} allows us to control the sparsity of the edge weights, we cannot, however, obtain graphs with a desired number of connected components. Therefore, we specialize \eqref{eq:PGL_modeling} to learn graphs factors with multiple connected components by solving the following \emph{rank-constrained product graph learning}~(\texttt{RPGL}) problem
\begin{align}
&\underset{\bbL_P \in \ccalL_P, \,\bbL_Q \in \ccalL_Q}{{\rm minimize}} \,
 {\rm tr}(\bbL_P \bbS_P) + {\rm tr}( \bbL_Q \bbS_Q) + h(\bbL_P,\bbL_Q) \nonumber \\
&\text{subject to} \quad {\rm tr}(\bbL_P) = P, \, {\rm tr}(\bbL_Q)= Q, \tag{$\mathcal{P}2$}\label{eq:RCPGL_modeling} \\
&\hskip14mm \quad {\rm rank}(\bbL_P) = R_P \quad \text{and} \quad {\rm rank}(\bbL_Q) = R_Q,
\nonumber
\end{align}
where $R_P = P - K_P$ and $R_Q = Q-K_Q$. Recall that the rank constraints ensure that the algebraic multiplicity of the zero eigenvalue of $\bbL_P$ and $\bbL_Q$ are, respectively, $K_P$ and $K_Q$. In~\cite{nie2016constrained}, a similar rank constraint was used to refine an affinity graph (with no product structure) for spectral clustering. In contrast to Problem~\eqref{eq:PGL_modeling}, the optimization problem \eqref{eq:RCPGL_modeling} is a nonconvex optimization problem. 
Although the trace equality constraints are not required anymore to avoid a trivial solution, we retain them here as it will be required when we solve Problem~\eqref{eq:RCPGL_modeling} using a cyclic minimization technique as discussed next.

Let us denote the cost function in \eqref{eq:RCPGL_modeling} as 
\[
 f(\bbL_P,\bbL_Q) = {\rm tr}(\bbL_P \bbS_P) + {\rm tr}( \bbL_Q \bbS_Q) + h(\bbL_P,\bbL_Q).
\]
As $\bbL_P \succeq {\bf 0}$ and $\bbL_Q \succeq {\bf 0}$, we can rewrite \eqref{eq:RCPGL_modeling} as
\begin{align}
&\underset{{\bbL_P \in \ccalL_P, \bbL_Q \in \ccalL_Q}}{{\rm minimize}} \,\,
 f(\bbL_P,\bbL_Q)  + \gamma_1 \sum_{i=1}^{K_P} \lambda_i(\bbL_P) + \gamma_2 \sum_{i=1}^{K_Q} \lambda_i(\bbL_Q) \nonumber \\
 &\quad \text{subject to} \quad \quad {\rm tr}(\bbL_P) = P \,\, \text{and} \,\, {\rm tr}(\bbL_Q) = Q, \label{eq:p2_1}
 \end{align}
where the tuning parameters $\gamma_1>0$ and $\gamma_2 > 0$ force the second and third terms of the objective to zero at optimality ensuring the rank of the optimal $\bbL_P$ and $\bbL_Q$ to be $R_P$ and $R_Q$, respectively. 
From \eqref{eq:KyFan_knorm}, Problem \eqref{eq:p2_1} will then be 
\begin{align}
&{{\rm minimize}} \,
 f(\bbL_P,\bbL_Q)  + \gamma_1 {\rm tr}(\bbV_P^T\bbL_P\bbV_P) + \gamma_2{\rm tr}(\bbV_Q^T\bbL_Q\bbV_Q) \nonumber 
 \\[0.5em]
 & \text{subject to} \quad  \bbL_P \in \ccalL_P, \, \bbL_Q \in \ccalL_Q, \, {\rm tr}(\bbL_P) = P, \, {\rm tr}(\bbL_Q) = Q \nonumber \\[0.2em]
 &\quad \quad \quad \quad \quad \bbV_P^T\bbV_P = \bbI_{K_P} \,\, \text{and} \,\, \bbV_Q^T\bbV_Q = \bbI_{K_Q} 
\label{eq:RCPGL_2}
\end{align}
with variables $\bbL_P \in \mathbb{S}_+^P$, $\bbL_Q \in \mathbb{S}_+^Q$, $\bbV_P \in \mathbb{R}^{P \times K_P}$, and 
$\bbV_Q \in \mathbb{R}^{Q \times K_Q}$. The optimization problem \eqref{eq:RCPGL_2} is not a convex optimization problem. Therefore, we propose to solve it by alternatingly minimizing it with respect to $\{\bbL_P, \bbL_Q\}$ and 
$\{\bbV_P,\bbV_Q\}$, while keeping the other variable fixed. For each subproblem, we achieve the global optimum.

\subsubsection{Update of $\{\bbL_P,\bbL_Q\}$}
 The second and third term in the objective function of \eqref{eq:RCPGL_2} \emph{regularize} the data with the low-dimensional embeddings as
\[
{\rm tr}(\bbL_i\bbS_i) + \gamma_i {\rm tr}(\bbV_i^T\bbL_i\bbV_i) = {\rm tr}\left(\bbL_i [\bbS_i + \gamma_i \bbV_i\bbV_i^T]\right). 
\]
Further, we have 
\[
{\rm tr}(\bbV_i^T\bbL_i\bbV_i) = {\rm vec}^T(\bbV_i\bbV_i^T) \bbD_i \bbl_i = \bbq_{{\rm v},i}^T\bbl_i
\]
for $i \in \{P,Q\}$. Let us define $\bbq^T_{\rm v} :=[\bbq_{{\rm v},P}^T, \bbq_{{\rm v},Q}^T]$ and $\bbq^T_{\rm r} :=[\bbq_{{\rm r},P}^T, \bbq_{{\rm r},Q}^T]$ with
$
\bbq_{{\rm r},i}^T = \bbq_{{\rm d},i}^T + \bbq_{{\rm v},i}^T
$
for $i \in \{P,Q\}$. The subscript ``${\rm r}$" denotes rank constraints. Then, for fixed $\bbV_P$ and $\bbV_Q$, we have the subproblem 
\begin{align}
&\underset{\bbl \in \mathbb{R}^M}{\rm minimize} \quad \frac{1}{2} \, \bbl^T {\rm diag}(\bbp_{\rm d}) \bbl \, + \, \bbq^T_{\rm r} \bbl \nonumber \\
& \text{subject to } \quad \bbC\bbl = \bbd \quad \text{and} \quad \bbl \succeq {\bf 0},
\label{eq:RCPGL_sub1}
\end{align}
where the constraints are the same as before. The solution to \eqref{eq:RCPGL_sub1} can be computed using Algorithm~\ref{alg:laplacian_estimator} as $\bbl^\star = \texttt{PGL}(\bbp_{\rm d}, \bbq_{\rm r}, \bbC, \bbd, \texttt{Tol}, \rho)$, or more efficiently (and equivalently) as $\bbl_P^\star$ and $\bbl_Q^\star$ can be computed seperately. 

\subsubsection{Update of $\{\bbV_P, \bbV_Q\}$}For fixed $\bbL_P$ and $\bbL_Q$, Problem \eqref{eq:RCPGL_2} reduces to an eigenvalue problem
\begin{align}
&\underset{\bbV_P, \bbV_Q}{{\rm minimize}} \,\,\,
 \gamma_1 {\rm tr}(\bbV_P^T\bbL_P\bbV_P) + \gamma_2{\rm tr}(\bbV_Q^T\bbL_Q\bbV_Q) \nonumber \\
 & \text{subject to} \quad \quad  \bbV_P^T\bbV_P = \bbI_{K_P} \,\, \text{and} \,\, \bbV_Q^T\bbV_Q = \bbI_{K_Q}. 
\label{eq:RCPGL_sub2}
\end{align}
The optimal solution of $\bbV_P$ and $\bbV_Q$, are, respectively, given by the eigenvectors of $\bbL_P$ and $\bbL_Q$ corresponding to the $K_P$ and $K_Q$ smallest eigenvalues. That is, $\bbV_P = \texttt{eigs}(\bbL_P, K_P) $ and $\bbV_Q = \texttt{eigs}(\bbL_Q, K_Q)$. Computing this partial eigendecomposition approximately costs $K_PP^2 + K_QQ^2$ flops~\cite{saul2000introduction}.

This cyclic minimization procedure is summarized in Algorithm~\ref{alg:laplacian_rank}. Each iteration of Algorithm~\ref{alg:laplacian_rank} incurs a computational complexity of about $(K_P+1)P^2 + (K_Q+1)Q^2$ flops. The graph factor Laplacian matrices $\bbL_P$ and $\bbL_Q$ from Algorithm~\ref{alg:laplacian_rank} that have $K_P$ and $K_Q$ connected components, respectively, can be used for clustering. The low-dimensional subspaces $\bbV_P$ and $\bbV_Q$ provide the low-dimensional embeddings of the nodes in $\ccalG_P$ and $\ccalG_Q$, respectively. These can be used to compute the low-dimensional embeddings of the nodes in $\ccalG_N$. Furthermore, the rows of $\bbV_P$ and $\bbV_Q$ can also be used to cluster the nodes in $\ccalG_P$ and $\ccalG_Q$ using standard spectral clustering algorithms such as $K$-means~\cite{ng2002spectral}. 

\begin{algorithm}[!t]
\caption{Rank-constrained product graph learning}\label{alg:laplacian_rank}
\begin{algorithmic}[1]
\Function{\texttt{RPGL}}{$\bbp$, $\bbq$, $\bbC$, $\bbd$, \texttt{Tol}, $\rho$, $\texttt{MaxIter}$, $K_P$, $K_Q$}

\State Initialize $k \gets {\bf 0}$ 
\While{ $k < \texttt{MaxIter}$} 
\State $\bbl \gets \texttt{PGL}(\bbp, \bbq, \bbC, \bbd, \texttt{Tol}, \rho)$
\Comment $\bbl = [\bbl_P^T,\bbl_Q^T]^T$
\State $\bbL_P \gets \texttt{mat}(\bbD_P\bbl_P,P,P)$
\State $\bbL_Q \gets \texttt{mat}(\bbD_Q\bbl_Q,Q,Q)$
\State $\bbV_P \gets \texttt{eigs}(\bbL_P, K_P)$
\State $\bbV_Q \gets \texttt{eigs}(\bbL_Q, K_Q)$
\State $k \gets k +1$
\EndWhile

\Return $\bbL_P$, $\bbL_Q$, $\bbV_P$, and $\bbV_Q$
\EndFunction
\end{algorithmic}
\end{algorithm}

\section{Approximating a graph by a product graph}\label{sec:factorization}

Suppose we have the Laplacian matrix of a Cartesian product graph $\bbL_N$ available or given an estimate of the Laplacian matrix from any of the state-of-the-art graph learnin methods, we may factorize it into its graph factors. This is useful for approximating large graphs by a Cartesian product of smaller graph factors; see the illustration in Fig.~\ref{fig:CartesianPG}(b). Approximating large graphs through a Cartesian product of smaller graph factors reduces storage and computational costs involved in many GSP tasks~\cite{sandryhaila2014big}. In this section, we assume that 
$\bbL_N$ is available or has already been computed from data and propose solvers to factorize $\bbL_N$ into its graph factors.

\subsection{Nearest Kronecker sum factorization}\label{sec:Kronsumfact}

To compute the Laplacian matrices $\bbL_P$ and $\bbL_Q$ from $\bbL_N$, we propose the following \emph{Kronecker sum factorization} (\texttt{KronFact}) problem
\begin{align}
&\underset{{\bbL_P \in \ccalL_P, \bbL_Q \in \ccalL_Q}}{{\rm minimize}} \quad
 \| \bbL_N - \bbL_P \oplus \bbL_Q\|_F^2 \nonumber \\
&\quad\text{subject to} \quad \quad {\rm tr}(\bbL_P) = P \,\, \text{and} \,\, {\rm tr}(\bbL_Q) = Q, \tag{$\mathcal{P}3$} \label{eq:GraphFactorize}
\end{align}
where the trace constraints are used to fix the scales of $\bbL_P$ and $\bbL_Q$. The constraint sets $\ccalL_P$ and $\ccalL_Q$ are the sets of all the valid combinatorial Laplacian matrices of size 
$P \times P$ and $Q \times Q$, respectively [cf. \eqref{eq:Lapconstraint} for the definition]. 

The optimization problem~\eqref{eq:GraphFactorize} can be equivalently expressed as the following convex optimization problem [see Appendix \ref{sec:QP_lapfac}]
\begin{align}
&\underset{\bbl \in \mathbb{R}^M}{\rm minimize} \quad \frac{1}{2} \, \bbl^T {\rm diag}(\bbp_{\rm f}) \bbl \, + \, \bbq^T_{\rm f} \bbl \nonumber \\
& \text{subject to } \quad \bbC\bbl = \bbd \quad \text{and} \quad \bbl \succeq {\bf 0}
\label{eq:QP1}
\end{align}
with known parameters ${\rm diag}(\bbp_{\rm f}) \in \mathbb{S}_{++}^M$, $\bbq_{\rm f} \in \mathbb{R}^M$, $\bbC \in~\mathbb{R}^{L \times M}$, and $\bbd \in \mathbb{R}^{L}$, where recall that $L = P+Q+2$. The subscript ``${\rm f}$" represents factorization. The parameters in the objective function are 
defined as
\begin{align}
{\rm diag}(\bbp_{\rm f}) :=& {\rm bdiag}[{\rm diag}(\bbp_{{\rm f},P}), \, {\rm diag}(\bbp_{{\rm f},Q})] \nonumber
\\ 
=& {\rm bdiag}[2Q\bbD_P^T\bbD_P, \, 2P\bbD_Q^T\bbD_Q]  
\label{eq:pf} 
\end{align}
and
\begin{align}
\bbq^T_{\rm f} :=& [\bbq^T_{{\rm f},P}, \bbq^T_{{\rm f},Q}] \nonumber 
\\ =& [-2 {\rm vec}^T({\bbI}_Q)\tbL_N\bbD_P, -2 {\rm vec}^T({\bbI}_P)\tbL_N^T\bbD_Q], \label{eq:q3}
\end{align}
where $\tbL_N \in \mathbb{R}^{Q^2 \times P^2}$ is the \emph{tilde transform} of $\bbL_N$ defined in \eqref{eq:maketilde1} and \eqref{eq:maketilde2}. The parameters related to the equality constraints are defined in \eqref{eq:C_and_d} [see Appendix~\ref{sec:lap_constraints}].

Although the need for the trace equality constraints is not evident in \eqref{eq:GraphFactorize}, it is easy to realize that without the trace equality constraints, \eqref{eq:QP1} will result in a trivial solution. Further, Problems \eqref{eq:QP0} and \eqref{eq:QP1} have similar form. The role of $\beta_1$ and $\beta_2$ in \eqref{eq:QP0} to control the sparsity and distribution of edge weights is now played by $P$ and $Q$, respectively. 

As seen before, the optimization problem~\eqref{eq:QP1} can be solved efficiently and optimally using Algorithm~\ref{alg:laplacian_estimator}. Specifically, the optimal solution to~\eqref{eq:GraphFactorize} is given as $\bbl^\star = \texttt{PGL}(\bbp_{\rm f}, \bbq_{\rm f}, \bbC, \bbd, \texttt{Tol}, \rho)$. It is computationally less expensive to solve for $\bbl_P$ and $\bbl_Q$ separately as earlier.

\subsection{Rank constrained nearest Kronecker sum factorization}\label{sec:Kronsumfact_rank}

To compute the graph factors $\ccalG_P$ and $\ccalG_Q$ with $K_P$ and $K_Q$ components, respectively, we propose to factorize the available $\bbL_N$ to estimate $\bbL_P$ and $\bbL_Q$ using \eqref{eq:GraphFactorize} with rank constraints as in \eqref{eq:RCPGL_modeling}. Specifically, we propose to solve the following \emph{rank-constrained Kronecker sum factorization} (\texttt{R-KronFact}) problem
\begin{align}
&\underset{{\bbL_P \in \ccalL_P, \bbL_Q \in \ccalL_Q}}{{\rm minimize}} \quad
 \| \bbL_N - \bbL_P \oplus \bbL_Q\|_F^2 \tag{$\mathcal{P}4$} \label{eq:GraphFactorize_rank} \\
 &\quad \text{subject to} \quad \quad {\rm tr}(\bbL_P) = P, \, {\rm tr}(\bbL_Q) = Q, \nonumber \\[0.2em]
 &\quad \quad \quad \quad \quad \quad \quad{\rm rank}(\bbL_P) = R_P \quad \text{and} \quad {\rm rank}(\bbL_Q) = R_Q, \nonumber
 \end{align}
where recall that $R_P = P - K_P$ and $R_Q = Q-K_Q$. 

Let us then define the cost function as
\[
g(\bbL_P,\bbL_Q) = \| \bbL_N - \bbL_P \oplus \bbL_Q\|_F^2. 
\]
Since $\bbL_P \succeq {\bf 0}$ and $\bbL_Q \succeq {\bf 0}$, we can equivalently write Problem~\eqref{eq:GraphFactorize_rank} as
\begin{align}
&\underset{{\bbL_P \in \ccalL_P, \bbL_Q \in \ccalL_Q}}{{\rm minimize}} \,\,
 g(\bbL_P,\bbL_Q)  + \gamma_1 \sum_{i=1}^{K_P} \lambda_i(\bbL_P) + \gamma_2 \sum_{i=1}^{K_Q} \lambda_i(\bbL_Q)
\label{eq:GraphFactorize_rank1}, \nonumber \\
 &\quad \text{subject to} \quad \quad {\rm tr}(\bbL_P) = P \,\, \text{and} \,\, {\rm tr}(\bbL_Q) = Q, 
 \end{align}
which for appropriately selected $\gamma_1>0$ and $\gamma_2 > 0$ will, respectively, make the second and third term of the objective zero at optimality. Thus the rank of the optimal 
$\bbL_P$ and $\bbL_Q$ will be $R_P$ and $R_Q$, respectively. From \eqref{eq:KyFan_knorm}, Problem \eqref{eq:GraphFactorize_rank1} can be written as
\begin{align}
&{{\rm minimize}} \,\,
 g(\bbL_P,\bbL_Q)  + \gamma_1 {\rm tr}(\bbV_P^T\bbL_P\bbV_P) + \gamma_2 {\rm tr}(\bbV_Q^T\bbL_Q\bbV_Q) \nonumber \\[0.2em]
 & \text{subject to} \quad \bbL_P \in \ccalL_P, \, \bbL_Q \in \ccalL_Q, \, {\rm tr}(\bbL_P) = P, \, {\rm tr}(\bbL_Q) = Q, \nonumber \\[0.2em]
 & \quad \quad \quad \quad \quad \bbV_P^T\bbV_P = \bbI_{K_P} \,\, \text{and} \,\, \bbV_Q^T\bbV_Q = \bbI_{K_Q} 
\label{eq:GraphFactorize_rank2}
\end{align}
with variables $\bbL_P \in \mathbb{S}_+^P$, $\bbL_Q \in \mathbb{S}_+^Q$, $\bbV_P \in \mathbb{R}^{P \times K_P}$, and 
$\bbV_Q \in \mathbb{R}^{Q \times K_Q}$. This optimization problem is not a convex optimization problem. Therefore, we propose to solve it by alternatingly minimizing it with respect to $\{\bbL_P, \bbL_Q\}$ and $\{\bbV_P,\bbV_Q\}$, while keeping the other variable fixed. As before, we obtain the global optimum for each subproblem.

\subsubsection{Update of $\{\bbL_P,\bbL_Q\}$}
From Section~\ref{sec:Kronsumfact} and using the facts that ${\rm tr}(\bbV_P^T\bbL_P\bbV_P) = {\rm vec}^T(\bbV_P\bbV_P^T)\bbD_P \bbl_P$ and ${\rm tr}(\bbV_Q^T\bbL_Q\bbV_Q) = {\rm vec}^T(\bbV_Q\bbV_Q^T)\bbD_Q \bbl_Q$, for fixed $\bbV_P$ and $\bbV_Q$, Problem \eqref{eq:GraphFactorize_rank2} simplifies to
\begin{align}
&\underset{\bbl \in \mathbb{R}^M}{\rm minimize} \quad \frac{1}{2} \, \bbl^T {\rm diag}(\bbp_{\rm f}) \bbl \, + \, \bbq^T_{\rm rf} \bbl \nonumber \\
& \text{subject to } \quad \bbC\bbl = \bbd \quad \text{and} \quad \bbl \succeq {\bf 0},
\label{eq:GraphFactorize_ranksub1}
\end{align}
where the parameter $\bbq_{\rm rf} = \bbq_{\rm f} + \bbq_{\rm v} \in \mathbb{R}^M$ with $\bbq_{\rm f}$ defined in~\eqref{eq:q3} and recall that 
\begin{align}
\label{eq:q1}
 \bbq_{\rm v}^T &= [ \bbq_{{\rm v},P}^T, \bbq_{{\rm v},Q}^T] \nonumber \\ 
 &= [\gamma_1{\rm vec}^T(\bbV_P\bbV_P^T)\bbD_P, \gamma_2 {\rm vec}^T(\bbV_Q\bbV_Q^T)\bbD_Q] \nonumber. 
\end{align}
The solution to \eqref{eq:GraphFactorize_ranksub1} can be computed using Algorithm~\ref{alg:laplacian_estimator} as $\bbl^\star = \texttt{PGL}(\bbp_{\rm f}, \bbq_{\rm rf}, \bbC, \bbd, \texttt{Tol}, \rho)$, or $\bbl_P$ and $\bbl_Q$ can be solved separately.
 
\subsubsection{Update of $\{\bbV_P, \bbV_Q\}$}
For fixed $\bbL_P$ and $\bbL_Q$, Problem~\eqref{eq:GraphFactorize_rank2} reduces to an eigenvalue problem \eqref{eq:RCPGL_sub2}. The optimal solution of $\bbV_P$ and $\bbV_Q$, are, respectively, given by the $K_P$ and $K_Q$ eigenvectors of $\bbL_P$ and $\bbL_Q$ corresponding the $K_P$ and $K_Q$ smallest eigenvalues. That is, $\bbV_P = \texttt{eigs}(\bbL_P, K_P) $ and $\bbV_Q = \texttt{eigs}(\bbL_Q, K_Q)$. 

To summarize, the proposed solution to \eqref{eq:GraphFactorize_rank} is the procedure $\texttt{RPGL}({\bbp_{\rm f}, \bbq_{\rm rf}, \bbC, \bbd, \texttt{Tol}, \rho, \texttt{MaxIter}, K_P, K_Q})$. The Kronecker sum factorization to compute of $\bbL_P$ and $\bbL_Q$ incurs about order $P^2 + Q^2$ flops. With rank constraints, each iteration of \texttt{R-KronFact} costs about order $(K_P+1)P^2 + (K_Q+1)Q^2$ flops.
\section{Numerical experiments} \label{sec:numerical_expt}

This section presents results from numerical experiments\footnote{Software and datasets required to reproduce the results from the paper are available at \url{https://github.com/SaiKiranKadambari/ProdGraphLearn}} to demonstrate the developed theory and evaluate the proposed product graph learning methods on synthetic and real datasets. Specifically, using synthetic data, we evaluate the performance of the proposed methods, namely, \texttt{PGL} and \texttt{KronFact} in terms of F-measure and the performance of \texttt{RPGL} and \texttt{R-KronFact} in terms of Normalized Mutual Information (NMI), and compare with state-of-the-art methods for graph learning and clustering. We then demonstrate the efficacy of these proposed algorithms on real datasets related to air quality index (AQI) monitoring on time-series collected at several Indian cities~\cite{aqidata} and to cluster multi-view object images from the COIL-20 image dataset~\cite{nane1996columbia}.

\begin{figure*}[ht]
	\centering
		\includegraphics[width=2\columnwidth]{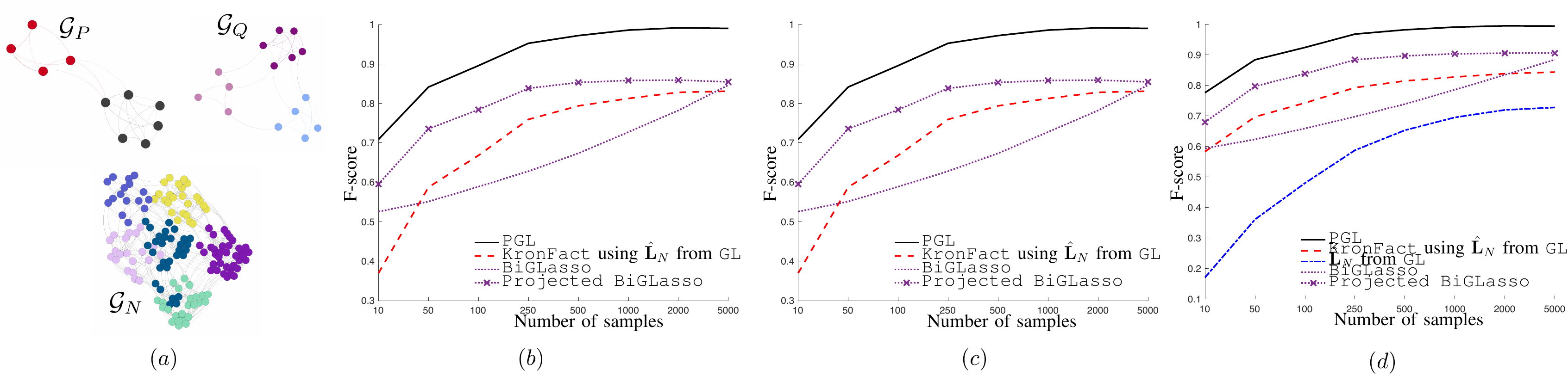}
	\caption{\small Product graph learning. (a) Synthetic graphs $\ccalG_P$ with $P=10$ nodes, $\ccalG_Q$ with $Q=15$ nodes, and $\ccalG_N$ with $N=150$ nodes. Graph learning performance: (b) Graph factor $\ccalG_P$. (c) Graph factor $\ccalG_Q$. (d) Product graph $\ccalG_N$.} 
	\label{fig:PGLsynthetic}
		
	\centering
		\includegraphics[width=2\columnwidth]{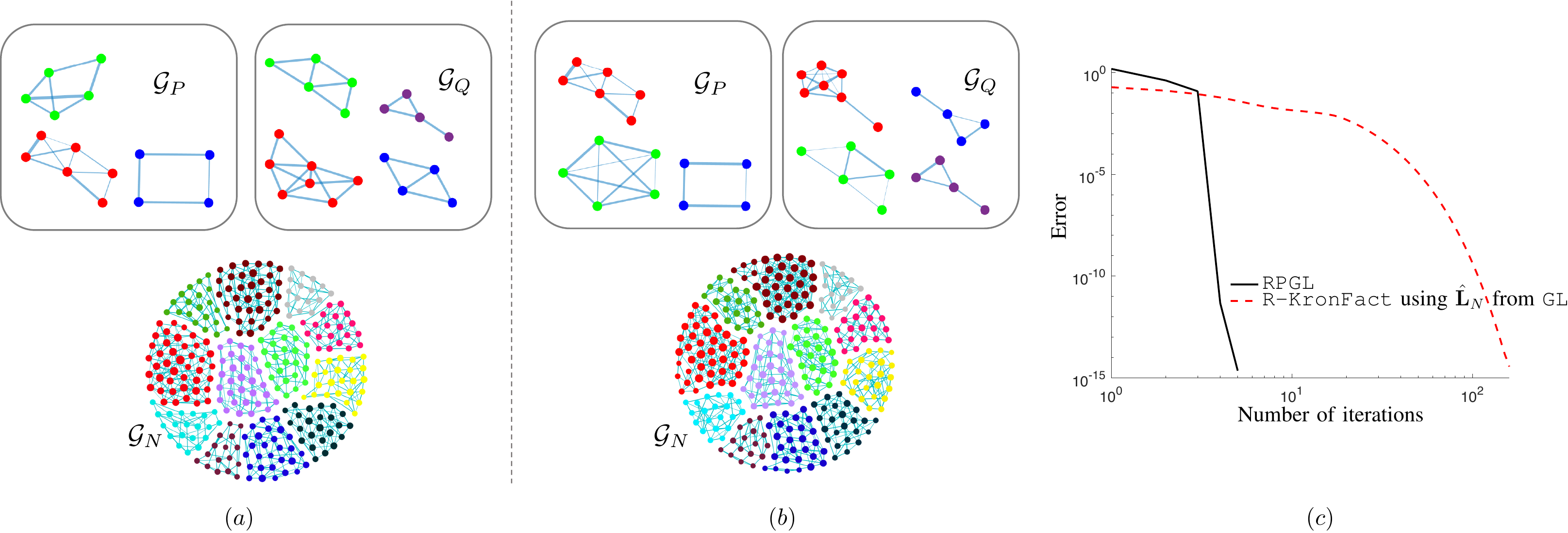}
	\caption{\small Product graph clustering. (a) Ground truth 3-component $\ccalG_P$ with $P=15$ nodes, 4-component $\ccalG_Q$ with $Q=20$ nodes, and 12-component $\ccalG_N$ with $N=300$ nodes. (b) Estimated multi-component graphs using \texttt{RPGL}. (c) Convergence of \texttt{RPGL} and \texttt{R-KronFact}.} 
	\label{fig:RPGLclusters}
\end{figure*}
\begin{table*}[ht]
\centering
 \begin{tabular}{ccccccccccccccccccccc}
\toprule
 &\multicolumn{7}{c}{Graph learning from $\bbX$}
&&
\multicolumn{3}{c}{\texttt{R-KronFact} using}\\
\cmidrule(r){2-8}\cmidrule(l){9-11}
& \texttt{PGL} & \texttt{RPGL}   & \texttt{SC}   & \texttt{$k$-means} & \texttt{Projected BiGLasso} & \texttt{GL} & \texttt{SGL}~\cite{kumar2019structured} & $\bbL_N$  &  $\hbL_N$ from \texttt{GL}~\cite{dong2016learning} & $\hbL_N$ from \texttt{SGL}~\cite{kumar2019structured} \\ 
\cmidrule(r){2-8}\cmidrule(l){9-11}
 $\bbL_P$ & 0.92& 0.92 & - & -& 0.79 & - & -& 1 & 0.25 & 1\\
 $\bbL_Q$ & 1& 1 & - & -& 0.80 & - & -& 1 & 0.71 & 0.98\\
  $\bbL_N$ & 0.95& {\bf 0.97} & 0.16 & 0.27& 0.61 & 0.44 & 0.94& {\bf 1} & 0.53 & {\bf 0.9}\\
\bottomrule
\end{tabular}
\caption{\small NMI for different graph learning methods. Since we aim to cluster the nodes in 
$\ccalG_N$, we do not report NMI values for methods that do not infer the graph factors. The NMI values for these cases are marked as ``-". The best result is highlighted in bold font.}
\label{tab:nmi}
\end{table*}

\subsection{Product graph learning (synthetic dataset)}\label{sec:synthetic_PGL}

To evaluate the performance of \texttt{PGL} and \texttt{KronFact}, we construct a graph $\ccalG_N$ by forming the Cartesian product of two community graphs $\ccalG_P$ and $\ccalG_Q$ as $\ccalG_N = \ccalG_P \diamond \ccalG_Q$. We use 
$P = 10$ and $Q = 15$ nodes so that $N = 150$. We choose the edge weights of the graph factors $\ccalG_P$ and $\ccalG_Q$ uniformly at random from the interval $[0.1,1]$. The synthetic graphs $\ccalG_P$, $\ccalG_Q$, and $\ccalG_N$ that we use
are shown in Fig.~\ref{fig:PGLsynthetic}(a).

We generate $T$ smooth graph signals $\{\bbx_i\}_{i=1}^T$ on the product graph $\ccalG_N$ using the Kronecker structured factor analysis model described in Section~\ref{sec:Kron_FA} with $\sigma^2=0$ (i.e., we have a noise-free setting).
Given these product graph signals, we estimate the graph factors $\ccalG_P$ and $\ccalG_Q$ using the following methods. (i) \texttt{PGL} that solves \eqref{eq:QP0}. (ii) \texttt{KronFact} that solves \eqref{eq:GraphFactorize} using estimated and noisy $\hat{\bbL}_N$ obtained by solving the convex program~\cite{dong2016learning}:
\begin{eqnarray}
&&\underset{\bbL_N \in \ccalL_N}{\rm minimize} \quad
 {\rm tr}(\bbL_N \bbS_N)  + \beta \|\bbL_N\|_F^2 \nonumber \\
&&\text{subject to} \,\,\,\,\,\, {\rm tr}(\bbL_N) = N, \label{eq:Lnestimate}
\end{eqnarray}
where $\bbS_N = \bbX\bbX^T \in \mathbb{R}^{N \times N}$ is the known sample data covariance matrix, and $\beta$ is the regularization parameter with which the sparsity of the edge weights may be controlled. We refer to the method that infers $\ccalG_N$ from \eqref{eq:Lnestimate} by ignoring the product graph structure as \emph{graph learning} (\texttt{GL}). (iii) \texttt{BiGLasso} from~\cite{kalaitzis2013Biglasso} that estimates precision matrices based on a probabilistic graphical data model. (iv) Since the precision matrices estimated from \texttt{BiGLasso} are not valid Laplacian matrices, we project them onto the set of all the valid Laplacian matrices by solving  
\begin{equation}\label{eq:ProjectionStep}
\begin{aligned}
\underset{{\bbL_i \in \ccalL_i}}{{\rm minimize}}
&& \|\bbP_i - \bbL_i\|_F^2 , \quad i \in \{{\rm P,Q}\}, 
\end{aligned}
\end{equation}
where $\bbP_P \in \mathbb{S}_+^{P}$ and $\bbP_Q \in \mathbb{S}_+^{Q}$ are the estimated precision matrices from \texttt{BiGLasso}. We refer to this method as \texttt{Projected BiGLasso}. While \texttt{GL} directly estimates $\ccalG_N$, the inferred graph factors from \texttt{PGL, KronFact, BiGLasso, Projected BiGLasso} are then used to construct the estimated product graph $\ccalG_N$. We use the following hyperparameters. For \texttt{PGL}, we use $\beta_1 = 0.2$, $\beta_2 = 0.3$, $\texttt{Tol}= 10^{-6}$, and $\rho =0.0051 $. For \texttt{GL}, we use $\beta = 1.5$.

We evaluate the performance of these Laplacian matrix estimators in terms of F-score, which is defined as
\[
\text{F-score}(\bbL,\hat{\bbL}) = \frac{2 \, {\rm tp}}{2 \, {\rm tp} + {\rm fn} + {\rm fp}},
\] 
where $\bbL$ is the ground truth Laplacian matrix, $\hat{\bbL}$ is the estimated Laplacian matrix, ${\rm tp}$ denotes true positive, {\rm fn} denotes false negative, and {\rm fp} denotes false positive. Although F-score is used to evaluate the performance of a binary classifier, it is a commonly used metric to evaluate graph learning algorithms as they too solve a binary hypothesis testing problem of inferring whether there exists a ``link" or ``no link" between a pair of nodes. An F-score of 1 means that all the edges have been perfectly inferred. 

In Figs.~\ref{fig:PGLsynthetic}(b),~\ref{fig:PGLsynthetic}(c), and~\ref{fig:PGLsynthetic}(d), we report for different sample sizes $T= \{10,50,250,500,1000,2000,5000\}$ the average F-scores (averaged over $10$ independent experiments) for inferring graphs $\ccalG_P$, 
$\ccalG_Q$, and $\ccalG_N$, respectively. As \texttt{PGL} exploits the underlying Cartesian product structure, the graph factor Laplacian matrices $\{\bbL_P,\bbL_Q\}$ can be inferred perfectly for a reasonable number of samples as compared to \texttt{GL}. We can see in Fig.~\ref{fig:PGLsynthetic}(d) that \texttt{KronFact}, which factorizes the estimated (noisy) $\bbL_N$ from \texttt{GL} to obtain the graph factor Laplacian matrices, has a better performance as \texttt{KronFact} accounts for the Cartesian product structure of $\ccalG_N$. The graph factor Laplacian matrices estimated from \texttt{Projected BiGLasso} have a better F-score as compared to \texttt{BiGLasso}, which estimates the precision matrices (interpreted as graph factor Laplacian matrices here). In essence, the proposed method outperforms the baseline methods for inferring graph factors and achieves a higher $\rm F$-score with a relatively smaller number of training samples.

\subsection{Product graph clustering (synthetic dataset)}

Next, we demonstrate the developed algorithms for clustering product graphs, which we perform via learning smaller multi-component graph factors from data. To do so, we generate a multi-component Cartesian product graph $\ccalG_N$ with $N = 300$ nodes that is formed by the Cartesian product of its graph factors $\ccalG_P$ and $\ccalG_Q$ having $P = 15$ and $Q = 20$ nodes, respectively. The graph factors $\ccalG_P$ and $\ccalG_Q$ have $K_P = 3$ and $K_Q = 4$ connected components, respectively. Thus the Cartesian product graph $\ccalG_N$ has $K = K_PK_Q = 12$ components. The graph factors and the Cartesian product graph are shown in Fig.~\ref{fig:RPGLclusters}(a). 

As before, we generate $T=1000$ graph signals on the product graph $\ccalG_N$ using the Kronecker-structured factor analysis model with $\sigma^2 =0$. Given these product signals, we estimate the multi-component graph factors using \texttt{RPGL} that solves \eqref{eq:RCPGL_modeling} using Algorithm~\ref{alg:laplacian_rank}, wherein we use the following hyperparameters for \texttt{RPGL}: $\beta_1= 0.25$, $\beta_2= 0.25$, $\texttt{Tol}=10^{-6}$, $\rho = 0.0051$, $\gamma_1= 0.5$, $\gamma_2=0.7$, $K_P=3$, $K_Q=4$, $\texttt{MaxIter}= 100$. The estimated graph factors $\ccalG_P$ and $\ccalG_Q$, and the multi-component product graph that is formed using $\ccalG_P$ and $\ccalG_Q$ are shown in Fig.~\ref{fig:RPGLclusters}(b), where we can see that the clusters are perfectly estimated. 

\begin{figure*}[t]	
  \centering
    \centering
    \includegraphics[width=1.25\columnwidth]{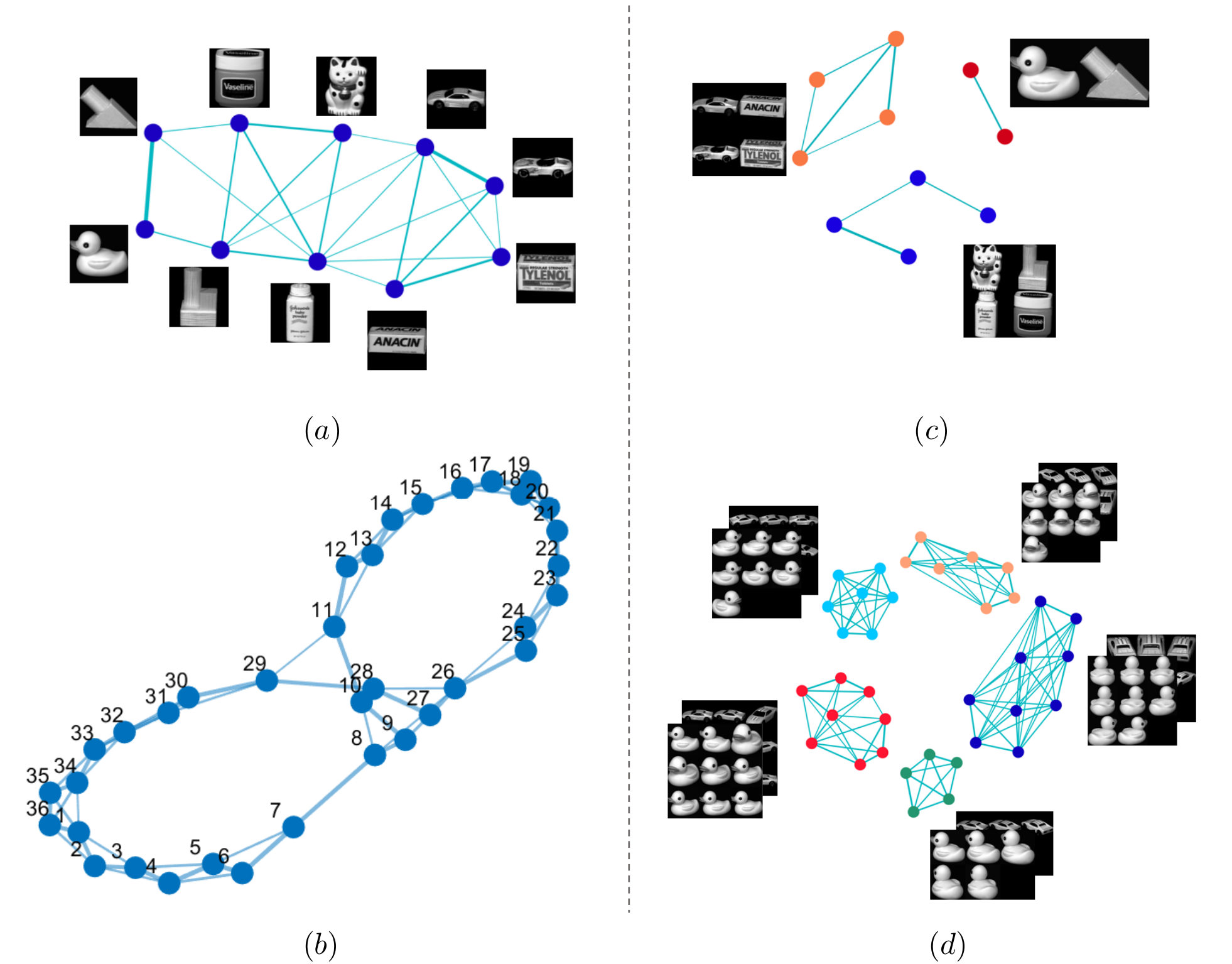}
    \caption{\small Product graph learning on multi-view COIL-20 dataset. Connected graph factors: (a) Estimated object graph 
    $\ccalG_P$. (b) Estimated view graph $\ccalG_Q$. Multi-component graph factors: (c) Estimated object graph $\ccalG_P$ with $K_P=3$ components. (d) Estimated view graph $\ccalG_Q$ with $K_Q = 5$ components.}
    \label{fig:coil}
   \vspace*{-4mm}
\end{figure*} 
Next, we compare the clustering accuracy to cluster nodes of $\ccalG_N$ in term of NMI, which is defined as
\[
\text{NMI}(\ccalC,\hat{\ccalC}) = \frac{I(\ccalC;\hat{\ccalC})}{0.5[H(\ccalC) + H(\hat{\ccalC})]},
\] 
where the set $\ccalC$ contains ground truth class assignments, the set $\hat{\ccalC}$ contains estimated clusters, $I$ is the mutual information between clusters $\ccalC$ and $\hat{\ccalC}$, and $H(\ccalC)$ and $H(\hat{\ccalC})$ represent the entropies. 
In Table~\ref{tab:nmi}, we provide NMI (averaged over 20 trials) based on cluster assignments estimated from the following methods. (i) Normalized spectral clustering (\texttt{SC})~\cite{ng2002spectral} and (ii) \texttt{$k$-means}~\cite{hastie2001statlearning} to compute the cluster assignments of the nodes in $\ccalG_N$, where \texttt{SC} constructs a similarity graph from data. We also use graph learning techniques: (iii) \texttt{GL} that estimates $\bbL_N$ by solving \eqref{eq:Lnestimate} and ignoring the product structure. (iv) Structured graph learning (\texttt{SGL})~\cite{kumar2019structured} that recovers a $K$-component graph Laplacian matrix ignoring the product structure. (v) Using \texttt{Projected BiGLasso} described in Section~\ref{sec:synthetic_PGL}. (vi) the proposed method \textemdash \texttt{PGL} that solves \eqref{eq:QP0}. (vii) the proposed method \textemdash \texttt{RPGL} that solves \eqref{eq:RCPGL_modeling} using Algorithm~\ref{alg:laplacian_rank}. (viii) the proposed method \textemdash \texttt{R-KronFact} that solves \eqref{eq:GraphFactorize_rank} based on the true graph $\bbL_N$ and estimated graphs $\hat{\bbL}_N$ from \texttt{GL} and \texttt{SGL}. For graph learning techniques that estimate $\bbL_P$, $\bbL_Q$, or $\bbL_N$ (for \texttt{PGL, RPGL}, and \texttt{R-KronFact} we construct $\bbL_N$ as $\bbL_P \oplus \bbL_Q$), we apply the $k$-means algorithm~\cite{hastie2001statlearning}, to estimate the cluster assignments and compute NMI, using the node embeddings $\texttt{eigs}(\bbL_P,K_P)$, $\texttt{eigs}(\bbL_Q,K_Q)$, and $\texttt{eigs}(\bbL_N,K_N)$. 

As can be seen in Table~\ref{tab:nmi}, \texttt{RPGL} results in the best NMI clustering accuracy and outperforms graph-learning based clustering technique \texttt{SGL}, where \texttt{RPGL} exploits the underlying product structure and performs clustering on large graphs by clustering its smaller graph factors. Using \texttt{R-KronFact}, approximating the available graph from \texttt{GL} or \texttt{SGL} by a Cartesian product graph, we perform better or on par with \texttt{GL} or \texttt{SGL} in terms NMI, but by processing smaller graphs.

In Fig.~\ref{fig:RPGLclusters}, we show the convergence of error for \texttt{RPGL} and \texttt{R-KronFact}, which use alternating minimization. The error is defined as
\begin{equation}
\label{eq:Error}
    \text{Error} = \frac{\|\bbL_P^{(k)} -\bbL_P^{(k-1)}\|_F^2}{ \| \bbL_P^{(k-1)} \|_F^2} + \frac{\|\bbL_Q^{(k)} -\bbL_Q^{(k-1)}\|_F^2}{ \| \bbL_Q^{(k-1)} \|_F^2},
\end{equation}
where $\bbL_i^{(k)}$ and $\bbL_i^{(k-1)}$, $i \in \{P,Q\}$ are the estimated Laplacian matrices at the current and previous iterations, respectively. As can be seen, \texttt{RPGL} that uses noise-free data as input converges in about 10 iterations, whereas \texttt{R-KronFact} that uses an estimate of the Laplacian matrix (can be interpreted as a noisy version of $\bbL_N$) from \eqref{eq:Lnestimate} as input needs about 100 iterations, for this dataset.

\subsection{Multi-view object clustering (COIL-20 dataset)}

In this section, we demonstrate product graph clustering using COIL-20 image dataset~\cite{nane1996columbia}. For illustration, we consider multi-view images of $10$ objects (selected at random from the available $20$ objects in the database). These objects were placed on a turntable, and we choose images of the objects taken from a fixed camera by rotating the table at an interval of 10 degrees. Thus the dataset has $360$ images corresponding to $10$ objects with $36$ views per image, where each image is of size $128 \times 128$ pixels. This corresponds to the multi-domain data with an object domain and a view domain.

This entire set of $360$ images can be efficiently represented as a product graph $\ccalG_N$ with $360$ nodes, which is approximated by the Cartesian product of the object graph $\ccalG_P$ with $10$ nodes and the view graph $\ccalG_Q$ with $36$ nodes, respectively. The $i$th pixel intensity corresponding to the 36 views of 10 objects collected in the data matrix $\bbX_i \in \mathbb{R}^{10 \times 36}$ forms the product graph signal. Since each image is $128 \times 128$ pixels, we have $T = 16,384$ features or samples. That is, the multi-domain data matrix $\bbX$ is of size $360 \times 16384$.

Given this multi-domain data that correspond to the objects and their views, the aim is to learn the Laplacian matrices $\bbL_P$ and $\bbL_Q$ associated with the graph factors $\ccalG_P$ and $\ccalG_Q$ that capture the dependencies between the objects and views, respectively. For this purpose, we use $\texttt{PGL}$ with hyperparameters $\beta_1 = 0.51$, $\beta_2 = 0.2$, $\texttt{Tol}=10^{-6}$, and $\rho = 0.0051$. The obtained graph factors corresponding to the objects and views are shown in Fig.~\ref{fig:coil}(a) and Fig.~\ref{fig:coil}(b), respectively. As we can see the objects with similar geometric shapes (e.g., \emph{Anacin} and \emph{Tylenol} images) are connected. The views having similar pose are connected (e.g., when the turntable is 10 degrees and 20 degrees). This intuitive result demonstrates that \texttt{PGL} captures similarities in the features, which can be leveraged to perform clustering as described next.

As we have seen before, we obtain multi-component graphs by imposing rank constraints on the Laplacian matrices. To cluster the objects and views from the multi-view COIL-20 dataset, we use \texttt{RPGL} to find multi-connected graph factors $\ccalG_P$ and $\ccalG_Q$ with $K_P = 3$ and $K_Q = 5$, respectively. This means that the object images are clustered into $3$ groups, and the views are clustered into $5$ groups. We use the following hyperparameters for \texttt{RPGL}: $\beta_1=0.5$, $\beta_2=0.1$, $\texttt{Tol}= 10^{-6}$, $\rho = 0.0051$, $\gamma_1=5$, $\gamma_2=10$, $K_P=3$, $K_Q=4$, $\texttt{MaxIter}=1000$.
 
In Fig.~\ref{fig:coil}(c), we show the inferred 3-component object graph, where the nodes are colored based on the cluster indices.
The objects with similar shapes are grouped to form 3 clusters. Similarly, in Fig.~\ref{fig:coil}(d), we show the inferred 5-component view graph, where the nodes are colored based on the cluster indices. As can be seen from images next to the graph, the views having similar poses are clustered together. 
%
%
\begin{figure}[t]
    \centering
    \includegraphics[width=0.7\columnwidth]{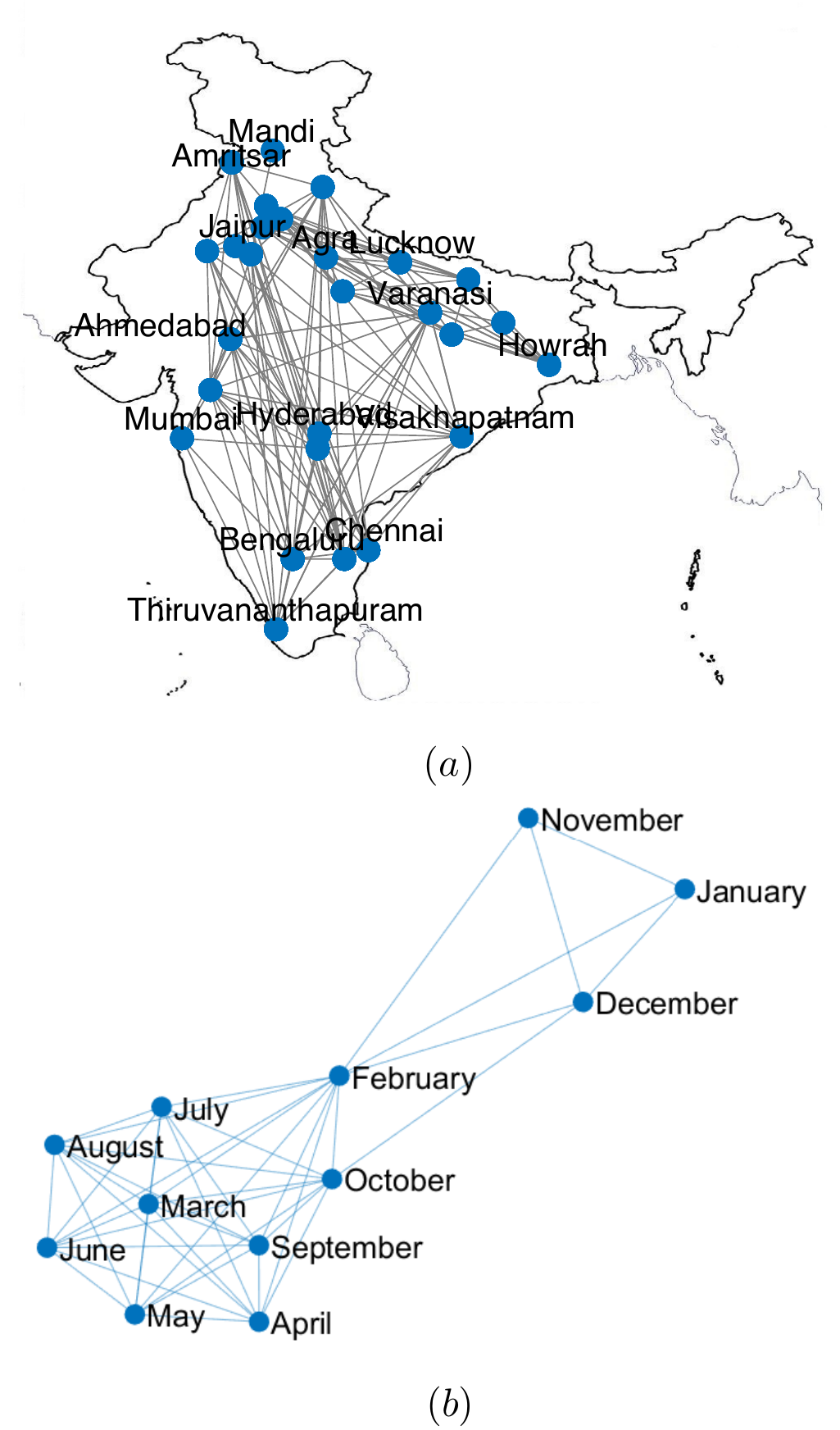}
        \caption{\small Product graph learning on air quality dataset. (a) Estimated spatial graph 
    $\ccalG_P$. (b) Estimated seasonal graph $\ccalG_Q$.}
    \label{fig:aqi}
    \vspace{-4mm}
\end{figure}
\subsection{Product graph learning with missing entries} \label{subsec:aqi}

Next, we apply the developed method for imputing missing ${\rm PM}_{2.5}$ measurements in the air quality data collected across $30$ air quality monitoring stations located all over India for the year $2018$~\cite{aqidata}. Specifically, at each air quality monitoring station, we have ${\rm PM}_{2.5}$ time-series corresponding to 360 days, but with many missing entries in the dataset. Given this multi-domain data with spatial and temporal domains, the aim is to learn a graph factor that captures the dependencies between the air quality measurements at different cities and a graph factor that captures the seasonal variations of the data. To do so, we view each space-time measurement as a signal indexed by the vertices of a product graph.

Let us collect the ${\rm PM}_{2.5}$ values from the air quality monitoring stations in $\bbX_i \in \mathbb{R}^{30 \times 12}$, $i=1,2,\ldots,T$ with $T=30$ samples per month. We split the available ${\rm PM}_{2.5}$ values into a training and test set. Let $\ccalA_i: \mathbb{R}^{30 \times 12} \rightarrow \mathbb{R}^{30 \times 12}$, $i=1,2,\ldots,T$ denote the mask that selects $85\%$ of the available measurements to form the training set. The observed product graph signals are collected in $\bbY_i = \ccalA_i(\bbX_i) \in \mathbb{R}^{30 \times 12}$, $i=1,2,\ldots,T$. This can be equivalently expressed as $\bby_i = \bbA_i\bbx_i$, where $\bbA_i \in \mathbb{R}^{360 \times 360}$ is the diagonal selection matrix related to $\ccalA_i$ and recall that $\bby_i={\rm vec}(\bbY_i)$ and $\bbx_i={\rm vec}(\bbX_i)$. Similarly, let $\ccalT_i: \mathbb{R}^{30 \times 12} \rightarrow \mathbb{R}^{30 \times 12}$, $i=1,2,\ldots,T$ denote the mask that selects the remaining $15\%$ of the available measurements to form the test set, which we use for validating the data imputation quality. The masks $\ccalA_i$ and $\ccalT_i$, $i=1,2,\ldots,T$ are known.

Since the observed data has many missing entries, we impute the missing entries in the data so that the imputed signal is smooth with respect to the product graph, which is not available. Therefore, we jointly solve the missing data imputation and product graph learning problem by solving the following optimization problem
\begin{align*}
&{\rm minimize} \quad \beta_1 \|\bbL_P\|_F^2+ \beta_2 \|\bbL_Q\|_F^2   \\
& \quad \quad +  \sum_{i=1}^{T} l(\bbX_i, \bbY_i) + \alpha_1 {\rm tr}( \bbX_i^T \bbL_P \bbX_i ) + \alpha_2{\rm tr}(\bbX_i \bbL_Q \bbX_i^T) \\
&\text{subject to} \quad \bbL_P \in \ccalL_P \quad \text{and} \quad \bbL_Q \in \ccalL_Q
\end{align*}
with variables $\left\{\bbX_i\right\}_{i = 1}^T$, $\bbL_P \in \mathbb{S}_+^{30}$, and $\bbL_Q \in \mathbb{S}_+^{12}$. Here, 
$
l(\bbX_i, \bbY_i) :=  \|\ccalA_i(\bbX_i - \bbY_i)\|_F^2 + \alpha_3 \|\bbX_i\|_F^2,
$
and $\alpha_1,\alpha_2$, and $\alpha_3$ are the regularization parameters. The loss function $l(\bbX_i, \bbY_i)$ is the Tikhonov-regularized least squares loss. The above optimization problem is nonconvex due to the coupling between the optimization variables in the objective function. Thus, we solve the problem by alternatingly minimizing it with respect to $\left\{\bbX_i\right\}_{i = 1}^T$ and $\{\bbL_P, \bbL_Q\}$, while keeping the other variable fixed. Given $\{\bbL_P, \bbL_Q\}$, we update $\left\{\bbX_i\right\}_{i = 1}^T$ by minimizing $\|\ccalA_i(\bbX_i - \bbY_i)\|_F^2 + \alpha_3 \|\bbX_i\|_F^2 + \alpha_1 {\rm tr}( \bbX_i^T \bbL_P \bbX_i ) + \alpha_2{\rm tr}(\bbX_i \bbL_Q \bbX_i^T)$ with respect to $\bbX_i$ for $i=1,2, \ldots, T$. From the first-order optimality condition, we have 
$
\ccalA_i(\bbX_i) + \alpha_1 \bbL_P \bbX_i + \alpha_2 \bbX_i\bbL_Q + \alpha_3 \bbX_i = \ccalA_i(\bbY_i),
$
which on vectorizing leads to 
$[\bbA_i + (\alpha_1 \bbL_P \oplus \alpha_2 \bbL_Q) + \alpha_3 \bbI_N]\bbx_i = \bbA_i\bby_i.$ Thus updating $\bbx_i = {\rm vec}(\bbX_i)$ can be done in closed form as
\[
\bbx_i = [\bbA_i + (\alpha_1 \bbL_P \oplus \alpha_2 \bbL_Q) + \alpha_3 \bbI_N]^{-1}\bby_i.
\]
Given $\left\{\bbX_i\right\}_{i = 1}^T$, we update $\{\bbL_P, \bbL_Q\}$ using \texttt{PGL} with hyperparameters $\beta_1 =5 $, $\beta_2 = 2$, $\texttt{Tol}=10^{-6}$, and $\rho = 0.51$. We use $\alpha_1 = 5.1 \times 10^{-4}$, $\alpha_2 = 10^{-4}$ and $\alpha_3 = 10^{-6}$, and perform these updates till
${\rm Error} \leq 10^{-3}$, where we recall the definition of ${\rm Error}$ from~\eqref{eq:Error}.

The inferred graph factor representing the similarity between the spatially distributed sensors is shown in Fig.~\ref{fig:aqi}(a). We observe from Fig.~\ref{fig:aqi}(a) that the links between the sensors are not based on the geometric proximity of the locations. The ${\rm PM}_{2.5}$ concentration is relatively higher during  November to February (winter months) compared to the other seasons. The graph capturing this seasonal variations for $12$ months of the ${\rm PM}_{2.5}$ data is shown in Fig.~\ref{fig:aqi}(b). We also evaluate the performance of the proposed graph learning method (PGL) in terms of the imputation error using the test set as $\frac{1}{T} \sum_{i = 1}^{T} \|\ccalT_i(\bbX_i - \hat{\bbX}_i)\|_F^2$, where $\hat{\bbX}_i$ is the estimated matrix without any missing entries. Using $k$-nearest neighbor graphs for $\bbL_P$ and $\bbL_Q$ with $k=10$ we get an imputation error of $0.04$, whereas the proposed joint graph learning and imputation method results in about an order of magnitude lower imputation error of $ 0.003$ on the test set. This also demonstrates that the proposed task-cognizant product graph learning (in this case, we learn the graph regularizer for the missing data imputation problem) outperforms approaches that construct graphs ignoring the task at hand. 

\section{Conclusions} \label{sec:conclusion}
We developed a framework for learning product graphs that can be factorized as the Cartesian product of two smaller graph factors. Assuming a smoothness data model, we presented efficient iterative algorithms to infer sparse product graphs. For product graph-based clustering, we presented an algorithm to infer multi-component product graphs and its low-dimensional representation by computing multi-component graph factors. We also presented a nearest Kronecker sum factorization algorithm to approximate sparse or multi-component graphs by sparse or multi-component Cartesian product graphs. With numerical experiments on real datasets, we demonstrated that the proposed algorithms that exploit the underlying Cartesian product structure have lower computational complexity and outperform the state-of-art graph learning methods.

\appendix

\subsection{Product graph Laplacian matrix constraints} \label{sec:lap_constraints}
The nonnegativity constraint on $\bbl$ is clear from its definition. The trace and null space equality constraints can be expressed as follows. Firstly, the trace equality constraint can be written as 
\begin{align*}
{\rm tr}(\bbL_P) &= {\rm vec}^T(\bbI_P) {\rm vec}(\bbL_P)  
 = {\rm vec}^T(\bbI_P) \bbD_P \bbl_P = P. 
\end{align*}
Next, the null space constraint can be written in the matrix-vector form as
\begin{align*}
{\bbL_P}{\bf 1} = ({\bf 1}^T \otimes \bbI_P) {\rm vec}(\bbL_P) 
= ({\bf 1}^T \otimes \bbI_P) \bbD_P\bbl_P = {\bf 0}.
\end{align*}
Stacking these two equations, we have
\[
\left[\begin{array}{c}{\rm vec}^T({\bbI}_P)\bbD_P \\({\bf 1}_P^T \otimes \bbI_P)\bbD_P\end{array}\right] \bbl_P  = \left[\begin{array}{c}P \\ {\bf 0} \end{array}\right] \Leftrightarrow \bbC_P \bbl_P = \bbd_P.
\]
Constructing  $\bbC_Q$ and $\bbd_Q$ along the similar lines, we can write the equality constraints of \eqref{eq:GraphFactorize}  as $\bbC \bbl = \bbd$, where $\bbC = {\rm bdiag}[\bbC_P, \bbC_Q]$ and $\bbd = [\bbd_P^T, \bbd_Q^T]^T.$

\vspace{-2mm}
\subsection{Efficient iterative solver for problems \eqref{eq:QP0} and \eqref{eq:QP1}} \label{sec:qpsolver}

Consider the following special case of a quadratic program (QP) with a diagonal matrix related to the quadratic term as
\begin{align}
&\underset{\bbl \in \mathbb{R}^M}{\rm minimize} \quad \frac{1}{2} \, \bbl^T {\rm diag}(\bbp) \bbl + \bbq^T \bbl \nonumber\\
& \text{subject to } \quad  \bbl \succeq {\bf 0} \quad \text{and} \quad \bbC\bbl = \bbd
\tag{{QP}}
\label{eq:templateQP}
\end{align}
with variable $\bbl \in \mathbb{R}^M$ and parameters ${\rm diag}(\bbp) \in \mathbb{R}^{M \times M}$, $\bbq \in \mathbb{R}^M$, $\bbC \in \mathbb{R}^{L \times M}$, and $\bbd \in \mathbb{R}^L$.  

By writing the Karush-Kuhn-Tucker (KKT) conditions and solving for $\bbl$ we obtain an explicit solution. The Lagrangian function for \eqref{eq:templateQP} is given~by
\begin{equation*}\label{eq:Lagrangian}
\ccalL(\bbl,\,\bblam,\,\bbmu) = \frac{1}{2} \bbl^T {\rm diag}(\bbp) \bbl + \bbq^T \bbl + \bbmu^T(\bbd -\bbC\bbl) -\bblam^T \bbl,
\end{equation*}
where $\bbmu \in \mathbb{R}^L$ and $\bblam \in \mathbb{R}^M$ are the Lagrange multipliers corresponding to the equality and inequality constraints, respectively. Then the KKT conditions are given by
\begin{equation*}
\begin{aligned}
&{\rm diag}(\bbp) \bbl^\star + \bbq - \bbC^T \bbmu^\star - \bblam^\star = 0, \\
\bbC \bbl^\star &= \bbd, \quad \bbl^\star \succeq {\bf 0},  \quad \bblam^\star \succeq 0, \quad \text{and} \quad \bblam^\star \odot \bbl^\star = {\bf 0}.
\end{aligned}
\end{equation*}
We next solve the KKT conditions to find $\bbl^\star$, $\bblam^\star$ and $\bbmu^\star$. To do so, first, we eliminate the slack variable $\bblam^\star$ 
and solve for $\bbl^\star$. This results in 
\[
\bbl^\star(\bbmu^\star) =\left\{{\rm diag}^{-1}(\bbp)[\bbC^T\bbmu^\star - \bbq] \right\}_+,
\]
where the projection $\{\cdot\}_+$ onto the nonnegative orthant is done elementwise by simply replacing each negative component of its argument with zero.

To find $\bbmu^\star$, we use $\bbl^\star(\bbmu^\star)$ in the second KKT condition. Specifically, we propose a simple iterative projected gradient descent algorithm to compute $\bbmu^\star$. The updates are given as
\begin{align}
& \bbl^{(k)}  =  \left\{{\rm diag}^{-1}(\bbp)[\bbC^T\bbmu^{(k)} - \bbq] \right\}_+,\\
&\bbmu^{(k+1)} = \bbmu^{(k)} - \rho [ \bbC\bbl^{(k)} - \bbd], 
\end{align}
where $\rho > 0$ is the step size. We initialize the iterations with $\bbmu^{(0)}$. The procedure is summarized as Algorithm~\ref{alg:laplacian_estimator}. The $\bbmu$ update step dominates the cost of computing $\bbl$. Although for 
a Laplacian matrix of size $N$, $M = N(N+1)/2$ and $L = N+1$, $\bbC$ is very sparse with $N(N+1)$ non-zero entries.
Thus the computational complexity of Algorithm~\ref{alg:laplacian_estimator} is approximately order $N^2$ flops.

\vspace*{-3mm}
\subsection{Expressing \eqref{eq:GraphFactorize} as \eqref{eq:QP1}} \label{sec:QP_lapfac}

The objective function of \eqref{eq:GraphFactorize} is given by
\begin{align}
\| \bbL_N - (\bbL_P \oplus \bbL_Q) \|_F^2 &= {\rm tr}(\bbL_N^2) + {\rm tr}( (\bbL_P \oplus \bbL_Q)^2) \nonumber \\ 
&\quad \quad \quad   -2 {\rm tr}(\bbL_N (\bbL_P \oplus \bbL_Q)).
\label{eq:objectivelapfac}
\end{align}

\subsubsection*{Second term}While the first term of \eqref{eq:objectivelapfac} does not affect Problem~\eqref{eq:GraphFactorize}, the second term can be further simplified~as
\[
{\rm tr}( (\bbL_P \oplus \bbL_Q)^2) 
 = Q {\rm tr}(\bbL_Q^2) + 2 {\rm tr}(\bbL_P) {\rm tr}(\bbL_Q) + P {\rm tr}(\bbL_P^2)(\bbL_Q)
\]
 Using the constraints ${\rm tr}(\bbL_P) = P$ and ${\rm tr}(\bbL_Q) = Q$, and from \eqref{eq:compact_LpLq}, we obtain 
\begin{align*}
{\rm tr}( (\bbL_P \oplus \bbL_Q)^2) 
 &= Q \bbl_P^T\bbD_P^T\bbD_P\bbl_P + P \bbl_Q^T\bbD_Q^T\bbD_Q\bbl_Q + 2PQ \\
 & =: \frac{1}{2} \bbl^T{\rm diag}(\bbp_{\rm f})\bbl^T + 2PQ,
\end{align*}
where recall that $\bbD_P^T\bbD_P$ and $\bbD_Q^T\bbD_Q$ are diagonal matrices. 

\subsubsection*{Tilde transform}To simplify the third term of \eqref{eq:objectivelapfac}, let us introduce the \emph{tilde transform} of a block matrix~\cite{koning1991block}. We can partition the product graph Laplacian matrix $\bbL_N$ as
\begin{equation}
\begin{aligned}
\label{eq:maketilde1}	
\bbL_N = \left[\begin{array}{ccc}\bbL_{11}  & \cdots & \bbL_{1Q} \\\bbL_{21}  & \cdots & \bbL_{2Q} \\\vdots  &  & \vdots \\ \bbL_{Q1}  & \cdots & \bbL_{QQ}\end{array}\right] \, \in \mathbb{R}^{QP \times QP},
\end{aligned}
\end{equation}
where each submatrix $\bbL_{mn}$ is of size $P \times P$. Then the tilde transform of such a block-partitioned matrix $\tbL_N$ is given as
\begin{equation}
\begin{aligned}
\label{eq:maketilde2}	
\tbL_N = \left[\begin{array}{c} 
{\rm vec}(\bbL_{11} )^T\\
{\rm vec}(\bbL_{21} )^T \\
\vdots \\
{\rm vec}(\bbL_{Q1} )^T \\
\vdots \\
{\rm vec}(\bbL_{1Q} )^T\\
{\rm vec}(\bbL_{2Q} )^T \\
\vdots \\
{\rm vec}(\bbL_{QQ} )^T 
\end{array}\right] \in \mathbb{R}^{Q^2\times P^2}.
\end{aligned}
\end{equation}

\subsubsection*{Third term} We can express ${\rm tr}(\bbL_N (\bbI_Q \otimes \bbL_P))$ as
\begin{align*}
{\rm tr}(\bbL_N (\bbI_Q \otimes \bbL_P)) &= {\rm vec}^T(\bbL_N) {\rm vec}(\bbI_Q \otimes \bbL_P) \\
&=  {\rm vec}^T(\tbL_N) [{\rm vec}(\bbI_Q) \otimes {\rm vec}(\bbL_P)] \\
& = {\rm vec}^T(\bbI_Q) \tbL_N {\rm vec}(\bbL_P) \\ &=  {\rm vec}^T(\bbI_Q) \tbL_N \bbD_P \bbl_P,
\end{align*}
where we use the tilde transform \eqref{eq:maketilde2} to arrive at the second equality. Similarly, we have
$
{\rm tr}(\bbL_N (\bbL_Q \otimes \bbI_P)) =  {\rm vec}^T(\bbI_P) \tbL_N^T \bbD_Q \bbl_Q.
$
Thus we have $-2 {\rm tr}(\bbL_N (\bbL_P \oplus \bbL_Q)) = \bbq_{\rm f}^T \bbl$ with
\[
\bbq^T_{\rm f} := [-2 {\rm vec}^T({\bbI}_Q)\tbL_N\bbD_P, -2 {\rm vec}^T({\bbI}_P)\tbL_N^T\bbD_Q].
\]

\bibliographystyle{IEEEtran}
\bibliography{sai20tsp}
\end{document}